\documentclass[final,5p,times,twocolumn]{elsarticle}

\usepackage[dvipsnames]{xcolor}
\usepackage{amssymb}
\newtheorem{definition}{Definition}
\newtheorem{example}{Example}
\usepackage[algo2e,ruled,noend,linesnumbered]{algorithm2e}
\usepackage{booktabs}
\usepackage{url}
\usepackage{amsmath}
\usepackage{subcaption}
\usepackage{float}
\usepackage{array}
\usepackage{ulem}
\usepackage{soul}
\usepackage{multirow}

\begin{document}

\begin{frontmatter}






\title{Smart Ride and Delivery Services with Electric Vehicles: Leveraging Bidirectional Charging for Profit Optimisation}




\author[1]{Jinchun Du\corref{cor1}}
\ead{jinchun.du@monash.edu}

\author[1]{Bojie Shen}
\ead{bojie.shen1@monash.edu}

\author[1]{Muhammad Aamir Cheema}
\ead{aamir.cheema@monash.edu}

\author[2]{Adel N. Toosi}
\ead{adel.toosi@unimelb.edu.au}

\cortext[cor1]{Corresponding author.}
\affiliation[1]{organization={Faculty of Information Technology, Monash University}, 
            city={Melbourne},
            postcode={3800}, 
            state={Victoria},
            country={Australia}}

\affiliation[2]{organization={School of Computing and Information Systems, University of Melbourne}, 
            city={Melbourne},
            postcode={3052}, 
            state={Victoria},
            country={Australia}}

\begin{abstract}
With the rising popularity of electric vehicles (EVs), modern service systems, such as ride-hailing delivery services, are increasingly integrating EVs into their operations. Unlike conventional vehicles, EVs often have a shorter driving range, necessitating careful consideration of charging when fulfilling requests. With recent advances in Vehicle-to-Grid (V2G) technology—allowing EVs to also discharge energy back to the grid—new opportunities and complexities emerge.
We introduce the Electric Vehicle Orienteering Problem with V2G (EVOP-V2G): a profit-maximization problem where EV drivers must select customer requests/orders while managing when and where to charge or discharge. This involves navigating dynamic electricity prices, charging station selection, and route constraints. We formulate the problem as a Mixed Integer Programming (MIP) model and propose two near-optimal metaheuristic algorithms: one evolutionary (EA) and the other based on large neighborhood search (LNS). Experiments on real-world data show our methods can double driver profits compared to baselines, while maintaining near-optimal performance on small instances and excellent scalability on larger ones. Our work highlights a promising path toward smarter, more profitable EV-based mobility systems that actively support the energy grid.

\end{abstract}


\end{frontmatter}


\section{Introduction}

Achieving net zero emissions is a global goal, with transportation being a major source of greenhouse gas (GHG) emissions.  
According to a 2022 analysis~\cite{rhg2024ghgemissions}, the transportation sector contributes about 16\% of global greenhouse gas emissions, with road transport making up approximately 70\% of that total.
Electric Vehicles (EVs) present a promising solution to mitigate these environmental impacts, leading to a projected decline in transport-related GHG emissions \cite{AustraliasEmissionsProjections2023}. The adoption of EVs also supports several United Nations Sustainable Development Goals (SDGs), particularly SDG 7 (Affordable and Clean Energy), SDG 11 (Sustainable Cities and Communities), and SDG 13 (Climate Action), by fostering clean energy use, sustainable mobility, and climate resilience~\cite{un_sdgs_2015}.

Beyond reducing emissions, EVs possess the potential to revolutionize the energy landscape through Vehicle-to-Grid (V2G) technology. 
V2G technology enables EVs to both store energy and transfer it back to the grid, homes, and other buildings. 
By enabling EVs to feed electricity back into the grid, V2G can optimize energy costs, enhance grid stability, and further contribute to emissions reduction. To fully harness the benefits of V2G and accelerate EV adoption, innovative approaches are required to seamlessly integrate travel and energy management for diverse EV users, including both private and commercial applications~\cite{cheema2024beyond}.

This research investigates profit maximization for single EV driver operating in commercial services like ride-hailing or delivery~\cite{shi2019operating,zalesak2021real}. Specifically, we focus on a variation of the Electric Vehicle Orienteering Problem (EVOP)~\cite{martins2021electric,karbowska2020genetic,wang2018electric,chen2020electric} where a set of potential orders, each with specified pickup and drop-off locations and time windows are given, and the EVOP seeks to determine the optimal subset of orders to accept, and the sequence in which to complete them. The objective is to maximize profit while considering factors such as travel distance, energy consumption, and charging infrastructure. The EV must be able to complete all accepted orders within their respective time windows without depleting its battery, and charging the EV if and when required.

Previous studies on the EVOP have largely focused on optimizing routes and energy consumption, without leveraging the potential of V2G technology to enhance profits. Although V2G presents significant opportunities for EV owners to increase revenue by purchasing energy at low prices and selling it back to the grid at higher prices, it also adds considerable complexity to the problem, which only a few previous works have investigated. Also, most existing studies not only have not considered the potential of V2G but also oversimplify the problem by assuming uniform charging rates and prices at all locations, which does not align with real-world conditions. For details, please refer to Section~\ref{sec::RelatedWork}.

\begin{figure}[bt]
    \includegraphics[width=\linewidth]{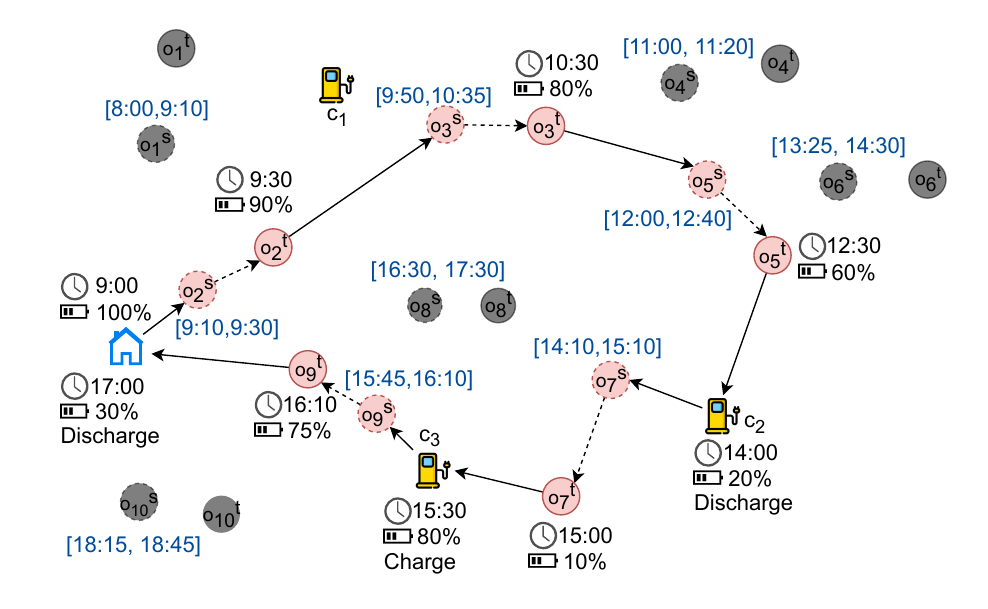}
    \caption{Representation of Electric Vehicle Orienteering Problem with V2G (EVOP-V2G) including 10 orders in order pool and 3 charging stations.}
    \label{fig::lns_fc}
\end{figure}

To the best of our knowledge, we are the first to formulate the Electric Vehicle Orienteering Problem with the integration of V2G technology to maximize profit. We call this the Electric Vehicle Orienteering Problem with V2G (EVOP-V2G). Fig.~\ref{fig::lns_fc} demonstrates a simple instance of the EVOP-V2G problem where an EV owner with workshift between 9:00am to 5:00pm has to maximise their profit. The order pool consists of 10 orders ($o_1$ to $o_{10}$) each with a time window (shown within square brackets) within which the order must be completed. For each order, the pick-up location ($p$) is represented by a dashed circle, while the drop-off location ($d$) is indicated by a solid circle. The battery icon represents the current state of charge (SoC) of the electric vehicle (EV), indicating the amount of energy available. The clock symbol denotes the specific time at which the EV arrives at a given location. The figure shows one possible schedule for the EV owner where it leaves home at 9:00am, completes $o_2$, $o_3$ and $o_5$ in this order before stopping at the charging station $c_2$ to discharge to 20\% SoC as there is additional energy. Then, it processes the order $o_7$ which depletes the battery to 10\%. It decides to visit charging station $c_3$ to charge to 80\% before proceeding to complete $o_9$. After completing $o_9$, with the end of the work shift approaching, it returns home. 
With bidirectional charging available at home, the EV discharges its battery down to a user-defined threshold (in this case, 30\%) when electricity prices are high, generating additional revenue.

While the example above shows one possible schedule, the EVOP-V2G problem requires finding the schedule that maximizes the profit. Finding a high-quality solution for EVOP-V2G is challenging because the algorithm must determine which subset of orders to serve and the sequence in which to complete them, while also considering charging and discharging between orders to maximize profit. Deciding which charging/discharging stations to use and when to charge or discharge, amidst different charging rates and prices, adds an additional layer of complexity.

To address these challenges, we first propose a Mixed Integer Programming (MIP) model that formulates the problem using mathematical equations. While the MIP model solves the problem optimally with off-the-shelf solvers, it suffers from poor scalability in large-scale instances. To mitigate this limitation, we design two suboptimal algorithms: Evolutionary Algorithm (EA) and Large Neighborhood Search (LNS).

The EA initializes a population pool containing a large number of feasible solutions and relies on bio-inspired processes to iteratively evolve these solutions towards better quality. Although simple, the main drawback of EA is its requirement to maintain a large population pool, leading to slow convergence in solution quality. Conversely, LNS, a local search-inspired algorithm, overcomes this issue by maintaining a single solution and systematically exploring a broader range of neighborhoods in the search space. We design an efficient LNS algorithm by designing advanced strategies to enhance its ability to effectively explore neighborhoods and quickly find high-quality solutions.

To evaluate our proposed algorithms, we derive order requests and charging station information directly from real-world data sources. We compare the three proposed algorithms with a greedy baseline approach, demonstrating that our algorithms can produce solutions with twice the profit. The results show that V2G contributes approximately 20\% of the total profit under our default settings. Among the three proposed algorithms, our MIP approach can optimally solve small-scale instances containing 30 orders and 3 charging stations, but it fails to scale for larger instances.  In contrast, EA and LNS achieve near-optimal solutions for small-scale instances and effectively scale for larger instances with 900 orders and 70 charging stations. LNS demonstrates faster convergence rates than EA, quickly finding initial solutions and rapidly improving them to high quality, outperforming EA in most instances
.

Our contributions in this paper are summarized below:
\begin{itemize}
    \item We formulate a novel optimization problem, EVOP-V2G, to maximize the profit of EV drivers in commercial settings such as ride-hailing. EVOP-V2G assumes a centralized server that broadcasts available orders for EV drivers, with each order associated with a monetary reward (e.g., ride fare). The objective of EVOP-V2G is to maximize the overall profit for the EV driver by optimizing both the selection of orders and the charging/discharging activities.

    \item 
    To solve EVOP-V2G, we introduce three methods: an optimal Mixed Integer Programming (MIP) model, and two suboptimal heuristic methods, Evolutionary Algorithm (EA) and Large Neighborhood Search (LNS). These methods effectively select customer orders and manage charging and discharging decisions.

    \item We conduct extensive experiments using a real-world dataset that includes realistic charging/discharging rates and prices, customer order requests, fare prices, and charging station information. Our findings highlight the effectiveness and efficiency of our proposed methods.

\end{itemize}

The structure of the paper is as follows: Section~\ref{sec::RelatedWork} reviews related work. Section~\ref{sec::preliminaries} formalizes the problem and describes its key aspects. Section~\ref{sec::Methodology} details our proposed methods: MIP, EA, and LNS. Sections \ref{sec::Evaluation} and \ref{sec::Conclusion} present the experimental results and conclusions, respectively.

\section{Related Work}
\label{sec::RelatedWork}


\subsection{Electric Vehicle Routing Problem}

The Vehicle Routing Problem (VRP), introduced in 1959~\cite{vrp}, is an optimization problem frequently encountered in logistics and has been extensively studied~\cite{lau2003vehicle, spliet2015time, lim2005multi, golden1984fleet, subramanian2012hybrid, stavropoulou2022consistent}. In VRP, a fleet of vehicles is tasked with serving a set of customers. The primary objective of the VRP is to determine optimal routes for each vehicle, aiming to minimize overall transportation costs or maximize operational efficiency. 
Recent VRP research has increasingly focused on Electric Vehicle Routing Problems (EVRPs), addressing the limited driving range of EVs and the need for recharging during operations.
Here, we provide a brief overview of relevant literature, considering the different variations of EVRP.

\begin{table*}[]
\setlength{\tabcolsep}{5pt} 
\centering
\footnotesize
\begin{tabular}{llcccccc}
\hline
\multirow{2}{*}{\textbf{Literature}} & \multirow{2}{*}{\textbf{Problem Type}} & \multicolumn{6}{c}{\textbf{Problem Characteristics}} \\ \cline{3-8} &   & \textbf{Single EV} & \textbf{Fleet of EV} & \textbf{Time Window} & \textbf{Charging}     & \textbf{Discharging} & \textbf{Time-variant prices} \\ \hline
Abdulaal et al.~\cite{abdulaal2016solving}   & EVRP   & \checkmark &  & \checkmark   & Full & \checkmark   & \checkmark \\ 
Narayanan et al.~\cite{narayanan2022reinforcement}   & EVRP & & \checkmark  &\checkmark   & N/A   & \checkmark   &  \checkmark  \\ 
Lin et al.~\cite{lin2021electric}& EVRP & & \checkmark  & \checkmark  & Partial      & \checkmark   &  \checkmark \\ \hline
Lee et al.~\cite{lee2013tour} & EVOP & \checkmark & & & Partial & &  \\
Karbowska-Chilinska and Chociej~\cite{karbowska2020genetic}                            & EVOP                          & \checkmark   &             &             & Full         &             &                      \\ 
Wang et al.~\cite{wang2018electric}  & EVOP  & \checkmark   &  & \checkmark   &Battery swap &   &     \\ 
Chen et al.~\cite{chen2020electric}                           & EVOP                          & \checkmark   &             & \checkmark   & Partial      &             &                      \\ 
Our work (EVOP-V2G) & EVOP & \checkmark & & \checkmark & Partial & \checkmark & \checkmark \\ \hline
\end{tabular}
\caption{Classification of the relevant literature based on problem characteristics closely related to our work (EVOP-V2G).}
\label{tab::related_work}
\end{table*}

\subsubsection{Single EV} 
Existing EVRP research has extensively studied this problem by exploring numerous variations (see \cite{kucukoglu2021electric} for a survey), however, the primary focus remains on fleets of EVs due to their critical role in urban logistics applications. In the context of a single EV in EVRP, the EV must depart from and return to the same depot, while visiting each customer exactly once. The primary objective of EVRP is to minimize energy consumption after servicing all customers, taking into account charging requirements. Several studies on single EVRP have focused on optimizing the routes of individual EVs while addressing different operational and logistical aspects such as non-linear charging~\cite{singleEV2} and the incorporation of multi-depot~\cite{abdulaal2016solving}.

\subsubsection{Fleets of EVs}
A fleet of EVs consist of multiple EVs, each of which requires departure and return to the same depot. While the EV fleet comprises more EVs, the objective of the EVRP remains same, i.e., the EVRP plans routes, one for each EV, to collectively meet the requirement of visiting each customer once while minimizing the energy consumption of all EVs.
Managing EV fleets models the practical urban logistics scenarios~\cite{hulagu2020electric,hiermann2016electric}, yet introduces additional complexities related to vehicle coordination and energy constraints. Fleets are generally categorized as homogeneous, where all vehicles share the same specifications~\cite{almouhanna2020location,ceselli2021branch}, or heterogeneous, where vehicles differ in parameters such as capacity or range~\cite{arias2015efficient,breunig2019electric}. The latter remains less explored due to its increased computational complexity.
In addition to that, EVRP studies have further examined diverse charging strategies, including full charging, where vehicles recharge to full capacity at charging stations~\cite{abdulaal2016solving,afroditi2014electric}; partial charging, which allows flexible energy replenishment~\cite{keskin2016partial,basso2021electric}; and battery swapping, where depleted batteries are exchanged for fully charged units~\cite{arias2015efficient,chen2016electric}. Time window constraints have also been considered~\cite{schneider2014electric}, requiring service at customers within predefined intervals and significantly increasing routing complexity.

\subsection{Electric Vehicle Orienteering Problem}

Differing from the VRP, where each customer must be visited exactly once,  the Orienteering Problem (OP) aims to maximize total profit by visiting a subset of customers within time or distance constraints. OP has been extensively studied~\cite{golden1987orienteering,fischetti1998solving,campos2014grasp,liang2013multiple}. A related variant, Vehicle Tour Planning (VTP), models tourist attractions instead of customers, assigning each a score to be maximized~\cite{ruiz2022systematic}.
Similar to EVRP, the Electric Vehicle Orienteering Problem (EVOP) has emerged recently which extends OP by incorporating EV-specific constraints, including charging needs, and aims to maximize profit by selecting an optimal subset of customers while accounting for energy limitations.

\subsubsection{Single EV}
For the single EV, EVOP assumes that the EV must both start and end its journey at the same depot. Each customer is associated with a profit value and can be distributed arbitrarily across the map. EVOP typically necessitates users to input constraints such as the time or distance that an EV can travel. By factoring in the locations of charging stations and the duration required for charging, EVOP selects a subset of customers that can maximize profit and plan an optimal route that satisfies the input constraint. Studies have explored various charging strategies in EVOP. For example, \cite{karbowska2020genetic} assumes full recharging at each station, while \cite{lee2013tour} incorporates waiting times before charging begins. Several EVOP variants introduce additional constraints, such as time windows~\cite{wang2018electric}, requiring visits within specified intervals, and range anxiety~\cite{chen2020electric}, which emphasizes maintaining a higher average state-of-charge due to limited charging infrastructure.

\subsubsection{Fleets of EVs}
For a fleet of EVs that contains multiple EVs, the objective of EVOP remains consistent, i.e., EVOP plans an individual route for each vehicle, aimed at maximizing the collected profit from each customer collectively. Each vehicle is required to depart from and return to the same depot, while ensuring that each customer is serviced by only one vehicle.
There is only limited number of studies focus on the EVOP problem for a fleet of EV~\cite{martins2021electric}. Additionally, the EVOP has been extended in various scenarios, including surveillance operations~\cite{mufalli2012simultaneous}, logistical operations in smart cities~\cite{panadero2020maximising,reyes2018team}, and the provisioning of charging services for geographically dispersed energy-critical sensors~\cite{xu2020approximation}. These scenarios underscore the necessity of managing limited resources effectively, often necessitating the deployment of a team or fleet of EVs and Unmanned Aerial Vehicles (UAVs) to fulfill task requirements.


\subsection{Discussion}

Our work differs significantly from prior studies in several important ways. First, with advances in battery technology and a growing focus on renewable energy, there is increasing interest in repurposing idle EV batteries for energy storage through V2G systems, which enable bidirectional energy flow between EVs and grid/building infrastructure. However, only a few EVRP studies have considered V2G~\cite{abdulaal2016solving, narayanan2022reinforcement, lin2021electric}, and even these are limited, e.g.,~\cite{narayanan2022reinforcement} considers only discharging, assuming EVs start with full batteries and don’t recharge. To our knowledge, V2G remains unexplored within the EVOP context. In this paper, we incorporate V2G into EV route planning to maximize user profit by efficiently serving diverse orders.
Second, while related to single-EV EVOP problems, our setting is more complex and general. We integrate charging and discharging as profit-affecting actions, with time-variant electricity prices influencing when and where these actions occur. Only a few EVRP studies address time-variant pricing~\cite{abdulaal2016solving,narayanan2022reinforcement,lin2021electric}, and none do so in EVOP. Additionally, we account for station-dependent charging rates, leading to varied charging durations and more nuanced decision-making.

To better differentiate our work from related studies, Table~\ref{tab::related_work} outlines the most closely related works to our problem, highlighting the key problem characteristics. Compared to related studies, our problem setting is more generalized where our work allows EVs to depart from and end at different locations, supports partial charging, incorporates dynamic pricing, leverages V2G, and includes time windows for orders.

\section{Preliminaries}
\label{sec::preliminaries}

\subsection{Problem Formulation}

To formulate the EVOP-V2G problem, we construct a directed query graph $G = (V_{c,o,s,d}, E, W)$. Here, $V_{c,o,s,d}$ denotes the combined set of vertices\footnote{In this paper, we often use $V_{x,y}$ to denote the union of two sets of vertices $V_x$ and $V_y$, i.e., $V_{x,y} = V_x \cup V_y$.}, comprising charging locations ($V_c$), service orders ($V_o$), as well as the source ($V_s$) and destination ($V_d$) locations for the EV driver. 
The edge set $E$ connects each vertex in $V_{c,o,s,d}$ to every other vertex to create a complete graph, i.e., $E = V_{c,o,s,d} \times V_{c,o,s,d}$. The weight function $w \in W$ assigns each edge $e_{ij} \in E$ a tuple $(d_{ij}, t_{ij})$, where $d_{ij}$ and $t_{ij}$ represent the travel distance and time from vertex $v_i$ to $v_j$, respectively.
The EV-specific information is often provided by the EV driver which includes 
the battery capacity $B$ (in kWh) and driving efficiency $\gamma$ (in kWh per km). Given the driving efficiency $\gamma$, we can compute the energy consumed along an edge with edge weight $(d_{ij},t_{ij})$ as $e = \gamma \times d_{ij}$. Our proposed techniques can easily integrate more sophisticated and accurate energy consumption estimation models if required. 

The services system contains a set of order $V_o$ that are available for the EV driver to choose from. Each order $v_i \in V_o$ is represented as a tuple $(l_i^p, l_i^d, p_i, d_i, t_i, [\tau_i^b, \tau_i^e])$, where $l_i^p$ represents the pick up location and $l_i^d$ represents the drop off location, $p_i$ denotes the profit earned for completing the order $v_i$, $d_i$ and $t_i$ represent the distance and time, respectively, of $v_i$ when traveling between $l_i^p$ and $l_i^d$, and $[\tau_i^b, \tau_i^e]$ indicates the time window within which an EV does not arrive earlier for pickup than $\tau_i^b$ and completes the drop-off no later than $\tau_i^e$. 
Note that, when calculating the travel distance/time from an order vertex $v_i$ to another vertex in the graph $G$, we consider the EV departing from the drop-off location $l_i^d$. Similarly, when computing the travel distance/time from a vertex to an order vertex $v_i$, we consider the EV reaching the pick-up location $l_i^p$.

For each charging station $v_i \in V_c$, 
\hbox{${P}_i$} is given to denote the charging/discharging rate in kW.
The planning horizon of charging/discharging is discretised into $|T|$ number of disjoint, consecutive timeslots, $T = \{t_0, \cdots, t_t\}$, each of which has the length $\delta$. 
We assume the charging/discharging price are represented as piece-wise constant function at each charging station $v_i$. $P^{Bt_{k}}_{i}$ denotes the charging price at $v_i$ (in \$ per kWh) at timeslot $t_k \in T$, while $P^{St_{k}}_{i}$ is the discharging price at $v_i$ (in \$ per kWh) at timeslot $t_k \in T$. 
By default, the planning horizon $T$ is set to 24 hours to accommodate EV charging/discharging at both the source and destination locations.
Table~\hbox{\ref{tab:notation_summary}} summarizes the notation used throughout this paper. Using this specified notation, we formulate the EVOP-V2G query as follows:

\begin{definition}
\textbf{EVOP-V2G Query}:
A query is defined as a tuple $Q = (V_s, V_d, B, [B^s, B^d], [\tau^b_w, \tau^e_w])$, where an EV driver starts from source location $V_s \in V$ with initial battery level $B^s$, and must reach destination $V_d \in V$ with at least $B^d$ battery remaining. The EV has a total battery capacity of $B$ and operates within a working hour window $[\tau^b_w, \tau^e_w]$, during which it may serve orders ($v_o \in V_o$) or charge/discharge at stations ($v_c \in V_c$). Outside this window, only charging/discharging at $V_s$ and $V_d$ is allowed. 

The output is a sequence of actions (serving orders or charging/discharging) aligned with a discretized planning horizon $T$. EVOP-V2G evaluates orders from the order pool $O$ and selects a subset that optimally balances route planning and charging/discharging schedules. The objective is to maximize the profit potential for EV drivers operating within predefined working hours $[\tau^{b}_w, \tau^{e}_w]$.

\end{definition}


\begin{table}[t]
\centering
\scriptsize
\caption{Summary of Notation}
\label{tab:notation_summary}
\begin{tabular}{cp{7cm}}
\toprule
\textbf{Symbol} & \textbf{Definition} \\
\midrule
$G$ & A complete and directed graph. \\
$E$ & A set of edges in $G$. \\ 
$V_o$ & A set of vertices each of which associate with an order.\\
$V_c$ & A set of vertices each of which associate with a charging station.\\
$V_s$ & The source location of EV.\\
$V_d$ & The destination location of EV.\\
$T$ & The planning horizon $T$ of charging/discharging at station. \\
$\delta$ & The length of each timeslots in the planning horizon $T$. \\
$B$ & The battery capacity of the EV.\\
$\gamma$ & The driving efficiency of the EV. \\
$P_i$ & The charging rate at charging station $v_i \in V_c$. \\
$P^{Ct_k}_{i}$ & The charging price at charging station $v_i \in V_c$ at time period $t_k$. \\
$P^{Dt_k}_{i}$ & The discharging price at charging station $v_i \in V_c$ at time period $t_k$. \\
$l_i^p$ & The pick up location of an order $v_i \in V_o$.\\
$l_i^d$ & The drop off location of an order $v_i \in V_o$.\\
$d_{i}$ & The travel distance between $l_i^p$ and $l_i^d$ when completing order $v_i \in V_o$. \\
$t_{i}$ & The travel time between $l_i^p$ and $l_i^d$ when completing order $v_i \in V_o$. \\
$p_i$ & The profit for an order $v_i \in V_o$. \\
$[\tau^b_{i}, \tau^e_{i}]$ & 
The time window for order $v_i \in V_o$ starts at $\tau^b_{i}$ and ends before $\tau^e_{i}$. \\
$[\tau_w^s, \tau_w^e]$ & The working hours of an EV driver starts at $\tau_w^s$ and ends before $\tau_w^e$. \\
$[B^s, B^d]$ & The initial battery level $B^s$ at source, and minimal battery required $B^d$ at destination.\\
\bottomrule
\end{tabular}
\end{table}


\section{Methodology}
\label{sec::Methodology}
In this section, we present three novel algorithms for solving the EVOP-V2G. Firstly, we formulate the problem as a Mixed-Integer Programming (MIP) model to achieve optimal solutions. However, MIP model suffers from computational burdens and struggles with scalability. To address this, we propose two suboptimal algorithms: the Evolutionary Algorithm (EA), a bio-inspired method that iteratively evolves solutions, and Large Neighborhood Search (LNS), a local search-based approach that efficiently explores the search space to find suboptimal solutions.


\begin{figure*}[ht!]
\begin{align}
    & 
       max \sum_{v_i \in V_{o}}\sum_{v_j \in V_{o,c,d}}[x_{ij}p_i] +  
        \sum_{v_i \in V_{c}}\sum_{t_k \in T}[dc^{t_k}_{i}P^{Dt_{k}}_{i} - rc^{t_k}_{i}P^{Ct_{k}}_{i}] 
    \label{eq:ob}\\
    s.t. &
    \sum_{v_i \in V_{o,c,d}} x_{ji} = 1, & v_j \in V_{s} \label{eq:2}\\
    & \sum_{v_i \in V_{o,c,s}} x_{ij} = 1, & v_j \in V_{d} \label{eq:3}\\
    & \sum_{v_i \in V_{o,c,s}} x_{ij} \leq 1, & \forall v_j \in V_{o} \label{eq:4}\\
    & \sum_{v_i \in V_{o,c,s}} x_{ij} - \sum_{v_i \in V_{o,c,d}} x_{ji}  = 0, & \forall v_j \in V_{o,c} \label{eq:5} \\
    & \tau_{i} = T[0] & v_i \in V_{s} \label{eq:5_add} \\
    & \tau_{i} = T[|T| - 1] & v_i \in V_{d} \label{eq:5_add2} \\
    & \tau^{b}_w \leq \tau_{i} \leq \tau^{e}_w & \forall v_i \in V_{o,c} \label{eq:6} \\
    & \tau^b_{i} \leq \tau_i \leq \tau^e_{i}, & \forall v_i \in V_{o} \label{eq:7}\\
    & \tau_{i}  + (t_{ij} + t_{i} )x_{ij} - M(1-x_{ij}) \leq \tau_{j} & \forall v_i \in V_{o,c,s}, \forall v_j \in V_{o,c,d} \label{eq:8}\\
    & (k+1)\delta(rc^{t_k}_{i} + dc^{t_k}_{i}) + t_{ij}x_{ij} - M(1-x_{ij}) \leq \tau_{j}  &  \forall v_i \in V_{c}, \forall v_j \in V_{o,c,d}, \forall t_k \in T \label{eq:9}\\
    & \tau_{i}  - k\delta \leq M(1 - rc^{t_k}_{i} - dc^{t_k}_{i}) &  \forall v_i \in V_{c}, \forall t_k \in T \label{eq:10}\\
    & (k+1)\delta(rc^{t_k}_{i} + dc^{t_k}_{i}) \leq \tau^d &  \forall v_i \in V_{c}, \forall t_k \in T \label{eq:11}\\
    & b_{i} = B^s , & v_i \in V_s  \label{eq:12}\\ 
    & b_{i}  \geq B^d , & v_i \in V_d \label{eq:13}\\
    & b_j \leq b_i - (\frac{(d_{ij} + d_i^s)\gamma}{\lambda})x_{ij} + M(1 - x_{ij}), & \forall v_i \in V_{o,s}, \forall v_j \in V_{o,c,d} \label{eq:14}\\
    & b_j \leq b_i + \sum_{t_k \in T}\delta rc^{t_k}_{i} - \sum_{t_k \in T}\delta dc^{t_k}_{i} - \frac{d_{ij}\gamma }{\lambda} x_{ij} + M(1 - x_{ij}), & \forall v_i \in V_c, \forall v_j \in V_{o,c,d}\label{eq:15} \\
    & rc^{t_k}_{i} + dc^{t_k}_{i} \leq 1  & \forall v_i \in V_{c}, \forall t_k \in T \label{eq:16}\\
    & \sum_{t \in T} \delta rc^t_{i} \leq B - b_i,   & \forall v_i \in V_{c} \label{eq:17}\\
    & \sum_{t \in T} \delta dc^t_{i} \leq b_i,   & \forall v_i \in V_{c} \label{eq:18}\\
    & 0 \leq b_j \leq B \sum_{v_i \in V_{o,c,s}}x_{ij}, & \forall v_j \in V_{c} \label{eq:19}\\ 
    & x_{ij} \in \{0,1\}, & \forall v_i \in V_{o,c,s}, \forall v_j \in V_{o,c,d} \label{eq:20}\\ 
    & rc^{t_k}_{i}, dc^{t_k}_{j} \in \{0,1\}, & \forall v_i \in V_{c}, t_k \in T \label{eq:21} 
\end{align}
\end{figure*}

\subsection{Mixed-Integer Programming (MIP) Model} \label{sec:MIP}
In this subsection, we introduce MIP model to find optimal solution for EVOP-V2G. Given the input graph $G$, we first introduce the binary and continue decision variable as follow: 


\begin{itemize}
    \item The binary decision variables are: \\
$x_{ij} = $ 
    $\left\{
    \begin{array}{cl}
    1        & {  \text{ if edge ($v_i$,$v_j$) $\in E$ is traveled by an EV, }} \\
    0        & {  \text{ otherwise,}}\\
    \end{array} \right. $\\
$rc^{t_k}_{i} = $ 
    $\left\{
    \begin{array}{cl}
    1        & {  \text{ if the EV is recharging at $v_i \in V_c$ at timeslot $t_k$ }} \\
    0        & {  \text{ otherwise,}}\\
    \end{array} \right. $\\
$dc^{t_k}_{i} = $ 
    $\left\{
    \begin{array}{cl}
    1        & {  \text{ if the EV is discharging at $v_i \in V_c$ at timeslot $t_k$ }} \\
    0        & {  \text{ otherwise,}}\\
    \end{array} \right. $\\


    \item The continuous decision variables are: 
\begin{itemize} 
    \item $\tau_i:$ arrival time of EV at node $v_i$, \\
    \item $b_i:$ remaining energy upon arrival to node $v_i$
\end{itemize}

\end{itemize}

The binary decision variable $x_{ij}$ indicates whether an EV travels between vertex $v_i$ and $v_j$ or not. 
When $v_j \in V_o$ is an order vertex, the binary decision variable $x_{ij} = 1$ indicates that the EV driver will serve the order $v_j$ at pick-up location $l^p_j$. Conversely, when $v_i \in V_o$, $x_{ij} = 1$ indicates that the EV driver will complete the order $v_i$ at drop-off location $l^d_i$.
Additionally, we restrict that charging and discharging of EVs cannot occur simultaneously at the same timeslot $t_k$, this constraint will be imposed later on. Given the binary and continuous decision variable, the objective function is defined as Equation (\ref{eq:ob}). The objective function is designed to maximize the total profit collected for EV drivers, as well as the benefits received from discharging energy back to the grid through V2G technology. This sum of profit is then subtracted by the costs associated with charging the EV at stations.
Next, we defines the constraints for our MIP model.




\subsubsection{Constraints of MIP model}

The Constraint (\ref{eq:2}) - (\ref{eq:21}) enumerate all the constraints integrated into the MIP model. These constraints are classified into three categories: (i) Node Visiting Constraints; (ii) Travel Time Constraints; and (iii) EV Battery Constraints. The constraints (\ref{eq:19}) - (\ref{eq:21}) indicate the variables types and limits. Next, we provide a brief discussion of the constraints in each category:

\textbf{Node Visiting Constraints:}
Constraints (\ref{eq:2}) to (\ref{eq:5}) enumerate node visiting constraints. Specifically, constraints (\ref{eq:2}) and (\ref{eq:3}) ensure that the EV traverses precisely one edge from the source vertex (or other vertices) to other vertices (or the destination vertex), thereby enforcing departure from the source and arrival at the destination locations.
Constraint (\ref{eq:4}) describes that each order in $V_o$ can be visited at most once, preventing the EV from revisiting the same order multiple times.
Constraint (\ref{eq:5}) ensures that if an EV travels to a vertex $v_j$ that is an order or charging station, it must leave afterward, thereby necessitating the difference between $x_{ij}$ and $x_{ji}$ to be 0. Collectively, constraints (\ref{eq:2}) to (\ref{eq:5}) confine an EV to traverse a complete path
from source $V_s$ to destination $V_d$, and services the order $v_o \in V_o$ or engages in charging/discharging at station $v_c \in V_c$ in between. 
These constraints together are essential to ensure the path is complete and free of any subtours.

Our MIP model permits multiple visits to each charging station, including those at the source and destination, by introducing additional dummy vertices\footnote{The dummy vertex $v_c'$ essentially replicates the vertex $v_c \in V_c$ in the graph $G$.} for each charging station to allow for multiple visits. However, this would significantly increase the computational cost, as demonstrated in the paper~\hbox{~\cite{zhang2018electric}}. Therefore, we create these dummy vertices only when necessary for specific instances that require them (further details are provided in Section 5).

\textbf{Travel Time Constraints:}
Constraints (\ref{eq:5_add}) to (\ref{eq:11}) covers the travel time constraints for each vertex visits. 
Specifically, constraints (\hbox{\ref{eq:5_add}}) and (\hbox{\ref{eq:5_add2}}) set the arrival times at the source $V_s$ and destination $V_d$ to $T[0]$ and $T[|T|-1]$, respectively, aligning with the starting time and ending time of planning horizon $T$. Constraint (\hbox{\ref{eq:6}}) ensures that an EV driver can visit orders $V_o$ and charging stations $V_c$ only within the working hours $[\tau^{b}_w, \tau^{e}_w]$. 
Constraint (\ref{eq:7}) verifies the arrival time at each order $v_i \in V_o$, ensuring that the EV driver can arrive at the pickup location $l^p_i$ within its time window $[\tau_i^b, \tau_i^e]$ and pick up the customer on time.
Constraint (\ref{eq:8}) guarantees that travel between vertex $v_i$ and $v_j$, considering the travel time of edge $(v_i,v_j)$, ensures that if vertex $v_i \in V_o$ is an order, the arrival time at $v_j$ allows the EV driver sufficient time to complete the order.
Note that the notation $M$ stands for big M notation~\cite{bigM_notation}.
The Big $M$ notation works by incorporating a large constant $M$ into constraints to activate or deactivate them based on the value of a binary variable, ensuring that the constraint is enforced only when desired, e.g., Constraint (\hbox{\ref{eq:8}}) only valid if the edge $(v_i,v_j)$ is travelled, i.e., $x_{ij} = 1$.

Similarly, Constraint (\ref{eq:9}) ensures that the EV driver has adequate time to complete charging/discharging at the charging station $v_i \in V_c$ before reaching the next vertex $v_j$. Therefore, the constraint verifies for each timestep $t_k \in T$ that if the EV completes its charging or discharging (i.e., $(k+1)\delta(rc^{t_k}_{i} + dc^{t_k}_{i})$), it always has sufficient time to travel to the next vertex $v_j$.
Constraint (\ref{eq:10}) confirms that charging/discharging at charging stations adheres to practical considerations, i.e., the EV driver cannot charge/discharge at a charging station before arrival.
Finally, Constraint (\ref{eq:11}) ensures that charging/discharging operations are completed before the working hours of an EV driver.
Recall that we include two redundant vertices in $V_c$ to represent charging stations at the source and destination. Therefore, Constraints (\ref{eq:6}) and (\ref{eq:11}) are relaxed for these two redundant vertices to allow charging/discharging outside the working hours.

\textbf{EV Battery Constraints:}
Constraints (\ref{eq:12}) to (\ref{eq:18}) specify battery constraints for the EV.
At the start of the day, constraint (\hbox{\ref{eq:12}}) initializes the battery level of the EV at $V_s$ to $B^s$, allowing the EV to charge or discharge before the working hours begin. Similarly, constraint (\hbox{\ref{eq:13}}) requires that the battery level at $V_d$ must meet a minimum threshold of $B^d$. The EV may arrive with a lower battery level during the day, but it must have at least $B^d$ at the end of the day.
Constraint (\ref{eq:14}) monitors the battery level of an EV when traveling between vertices $v_i$ and $v_j$, ensuring that the battery level decreases according to the travel distance $d_{ij}$. Additionally, if $v_i$ is an order vertex, the constraint ensures that the battery level considers the travel distance $d^s_l$ when serving the order $v_i$.
Similarly, constraint (\ref{eq:15}) considers the EV charging $rc^{t_k}_{i}$ (resp. discharging $dc^{t_k}_{i}$) at the charging station $v_i \in V_c$ and ensures the battery level increase (resp. decrease) accordingly. 
Constraints (\ref{eq:16}) to (\ref{eq:18}) restrict the charging and discharging operations at a charging station $v_i \in V_c$. Constraint (\ref{eq:16}) ensures that when an EV arrives at $v_i$, charging and discharging cannot occur simultaneously. When the EV starts charging, constraint (\ref{eq:17}) ensures that the battery level of the EV does not exceed the battery capacity $B$. Conversely, when the EV starts discharging, constraint (\ref{eq:18}) ensures that the battery level of the EV decreases by at most the current battery level $b_i$. when it reaches $v_i$.

\begin{figure*}[bt]
    \includegraphics[width=\linewidth]{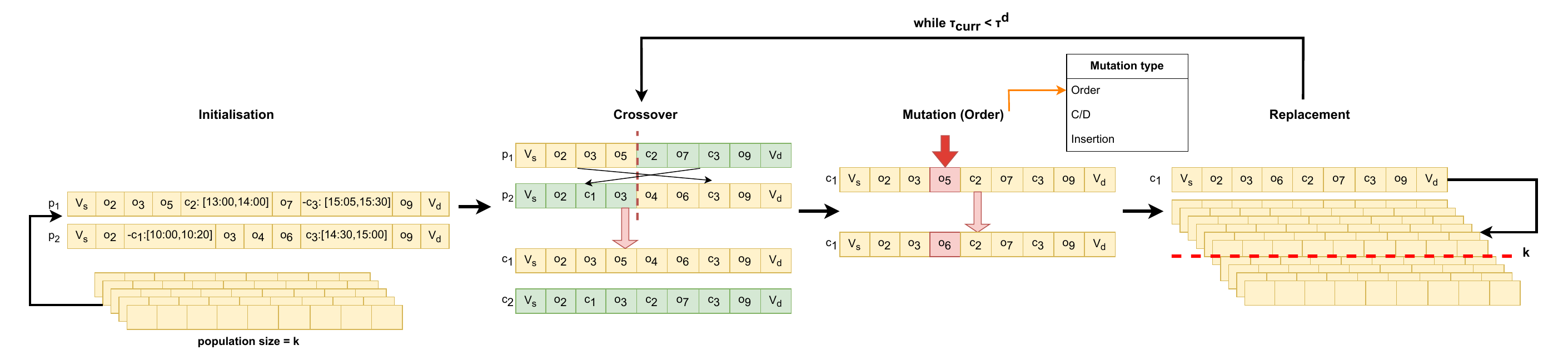}
    \caption{Illustration of the EA process, showing the stages of initialization, crossover, mutation, and replacement.
    }
    \label{fig::ga_fc}
\end{figure*}

\subsection{Evolutionary Algorithm}

Similar to the NP-hard EVRP and EVOP problems~\cite{kucukoglu2021electric}, our problem becomes more complex with the inclusion of V2G capabilities and dynamic charging/discharging prices. While MIP models can be solved by off-the-shelf solvers like Gurobi or CPLEX, they face two main limitations:
(i) MIP models are computationally intensive and do not scale well, as shown in our experiments; (ii) MIP models maintain the arrival time at each vertex, thus restricting visits to each charging station to only once by default. However, in certain scenarios, revisiting a charging station can be beneficial, particularly if it offers the lowest charging cost. In our experiments, we observed that enabling a single dummy vertex for a charging station can, in some cases, exponentially increase the computational time.


To address these issues, we explore Evolutionary Algorithms (EA), a meta-heuristic inspired by natural selection, that iteratively evolves feasible solutions. EA has been successfully applied to various combinatorial problems, including bin packing~\cite{binPacking_1,binPacking_2}, job-shop scheduling~\cite{jobScheduling_1,jobScheduling_2}, and EVRP~\cite{wang2023dual,zhu2021adaptive,zhenfeng2017electric}.
Following the success of EA, we propose EA-based algorithm in order to efficiently solve EVOP-V2G. Specifically, EA defines two important components: (i) \emph{Chromosomes}, which maintains the feasible solution of the problem. (ii) \emph{Fitness Function}, which evaluate the solution quality of the chromosomes. Fig.~\ref{fig::ga_fc} illustrates the process of the EA, providing an overview of its key steps and mechanisms. In board stoke, EA can be outlined as follows:

\begin{enumerate}
    \item \textbf{Initialisation}:
    Given the \emph{Chromosomes} and \emph{Fitness Function} defined, the initialization phase generates a pool of chromosomes, known as the population, which are then evaluated using the fitness function.
    
    
    \item \textbf{Selection \& Crossover}: 
    The selection \& crossover process simulates genetic exchange. Initially, the selection process chooses a pair of chromosomes from the population. Subsequently, the crossover selects a portion of the chromosome from each parent and exchanges them to produce new solutions, called offspring. The feasible offspring is then inserted back into the population pool.


    

    \item \textbf{Mutation}: For the chromosomes in the population, the mutation simulate genetic mutation by introducing random changes.

    \item \textbf{Replacement}: The steps 2 - 3 repeats until the chromosomes in the population have undergone crossover and mutation (given the probability). The replacement process maintains the population size by removing chromosomes with lower fitness values, ensuring that the population remains at the predefined size $k$ for the next generation.
        
    
\end{enumerate}

Steps 2 - 4 often refer to the evolution phases. Generally, EA requires maintaining a large number of chromosomes in the population pool, typically ranging from hundreds to thousands, depending on the application needs. To solve a proposed problem,  EA can run evolution phases for a specified number of iterations or terminate based on a predefined cut-off time. The best solution in the population is then returned as the result.


\subsubsection{Initialisation:}

For EA, an important phase involves designing an appropriate \emph{Chromosome} 
to represent a feasible solution of EVOP-V2G.
To formulate a EA-based algorithm for EVOP-V2G, we establish the chromosome as follows:

\begin{definition}
    \textbf{Chromosomes:} Given the input graph of EVOP-V2G, we represent chromosome $C$ as a sequence of actions, starting with $V_s$ and ending with $V_d$, categorized as follows: (i) Order-serving actions: These consist of a single vertex $(v_i)$, where $v_i \in V_o$. Each order must be served within its time window $[\tau^b_{i}, \tau^e_{i}]$, so we don't track specific arrival times. (ii) Charging/Discharging actions: Represented as $(v_i, [\tau_{c_{b}}, \tau_{c_{e}}])$, where $v_i$ is a charging station. A positive $v_i$ indicates charging, while a negative $v_i$ indicates discharging. $\tau_{c_{b}}$ and $\tau_{c_{e}}$ mark the start and end times of charging/discharging.

    
\end{definition}

Verifying the feasibility of the chromosome is straightforward and will be discussed in the later steps. As for the \emph{Fitness Function}, similar to the MIP models, we consider the charging/discharging price is discretized into timestep $t_k \in T$ and calculate the amounts of profits gained by the chromosome as follows:


\begin{equation}\label{eq:fitness}
f(C) = \sum_{\substack{(v_i) \in C}} p_i + \sum_{\substack{(v_i, [\tau_{c_{b}}, \tau_{c_{e}}]) \in C}} (\sum_{ \substack{t_k \in [\tau_{c_{b}}, \tau_{c_{e}}] \\ v_i < 0  }} P^{Dt_{k}}_{i} - \sum_{ \substack{t_k \in [\tau_{c_{b}}, \tau_{c_{e}}] \\ v_i > 0 }} P^{Ct_{k}}_{i} )
\end{equation}

\begin{algorithm2e}[t]
\small
\caption{Generating Chromosome}
    \label{algo:ga}
\setcounter{AlgoLine}{0}
    \SetKwInput{Input}{Input}
    \SetKwInput{Initialization}{Initialization}
    \SetKwInput{Output}{Output}
    \Input{[$\tau_w^s$, $\tau_w^d$]: EV driver's working hours, starting at $\tau_w^s$ and ending at $\tau_w^d$. [$B^s$, $B^d$]: Initial battery level $B^s$ at source, and minimum battery required $B^d$ at destination.}
    \Initialization{$\tau_{curr} = \tau^s$, $B_{curr} = B^s$, $chromosome = \emptyset$ }
    
    $V_o' = $ retrieve orders from $V_o$ within working hour [$\tau_w^s$, $\tau_w^d$]; \label{line:retrieve} \\
    append action $(V_s)$ to $C$ \\
    \While{$\tau_{curr} < \tau^d$ \label{line:whileloop}}
    {
        $v_i =$ randomly select an order from $V_o'$; \label{line:select_order}\\
        \If{$v_i$ is feasible under $(\tau_{curr}, B_{curr})$ \label{line:feasiable}}{
           append action $(v_i)$ to $C$ and update $(\tau_{curr}, B_{curr})$; \label{line:append_order}\\
        }
        remove order $v_i$ from $V_o'$; \label{line:remove}\\
    }
    append action $(V_d)$ to $C$ \label{line:appendD}  \\
    $\tau_{curr} = \tau^s$, $B_{curr} = B^s$; \label{line:reset} \\
    \For{each pair of order $(v_i,v_j) \in C$ \label{line:iterate_order}}{
        update $(\tau_{curr}, B_{curr})$ according to order $v_i$; \label{line:update_order}\\
        \For{i = 0 to randomNumber(0, m)-1\label{line:gen_random}}{
            $(v_c, [\tau_{c_{b}}, \tau_{c_{e}}])$ =  generate charging/discharging action; \label{line:gen_charging}\\
            \If{$(v_c, [\tau_{c_{b}}, \tau_{c_{e}}])$ is feasible under $(\tau_{curr}, B_{curr})$ \label{line:valid_charging} }{
                insert action $(v_c, [\tau_{c_{b}}, \tau_{c_{e}}])$ to $C$ 
                and update $(\tau_{curr}, B_{curr})$;\label{line:insert_update_charging}\\
            }
        }
    }
    \textbf{return} $C$; \label{line:return}
\end{algorithm2e}

\textbf{Generating Chromosome.} 
To initialize the EA population of size $k$, we propose a two-phase greedy algorithm that constructs chromosomes by first selecting order-serving actions from $V_o$ and then inserting up to $m$ random charging/discharging actions between consecutive orders.
The pseudo-code is provided in Algorithm~\ref{algo:ga}.
Given an EV’s working hours $[\tau_w^s, \tau_w^d]$ and battery constraints $[B^s, B^d]$, the algorithm begins by initializing an empty chromosome $C$, with the current time set to $\tau_{curr} = \tau_w^s$ and the battery level to $B_{curr} = B^s$.

\begin{itemize}
    \item Phase 1 (lines~\ref{line:retrieve}–\ref{line:appendD}): The algorithm starts with appending the source vertex $V_s$ to $C$. It then finds feasible orders that can be served within the working hours and iteratively appends randomly selected orders to $C$, ensuring time and battery constraints are respected. After each insertion, $\tau_{curr}$ and $B_{curr}$ are updated accordingly. This phase concludes by appending the destination vertex $V_d$.

    \item Phase 2 (lines~\ref{line:reset}–\ref{line:insert_update_charging}): The algorithm resets $\tau_{curr}$ and $B_{curr}$ and examines each consecutive pair of actions in $C$. Between these actions, it attempts to insert up to $m$ charging or discharging actions. For each insertion, a random charging/discharging action and a corresponding time interval $[\tau_{c_b}, \tau_{c_e}]$ are selected. Charging stations are then ranked based on a cost function:

        \begin{equation}
       \sum_{ \substack{t_k \in [\tau_{c_{b}}, \tau_{c_{e}}] \\ v_c < 0  }} P^{Dt_{k}}_{i} - \sum_{ \substack{t_k \in [\tau_{c_{b}}, \tau_{c_{e}}] \\ v_c > 0 }} P^{Ct_{k}}_{i} - p_e(v_{curr},v_c) - p_e(v_c,v_j)
        \end{equation}
    The current EV location is denoted by $v_{curr}$, and $v_c \in V_c$ represents a charging station. We rank the charging stations based on the charging (i.e., $v_c > 0$) or discharging (i.e., $v_c < 0$) price minus the energy cost from $v_{curr}$ to $v_c$ and from $v_c$ to $v_j$. 
    From the top five ranked charging stations, we randomly select one and add to the chromosome if timing and battery constraints are met. After processing all pairs, the chromosome $C$ is returned (line~\ref{line:return}); this process is repeated to populate the EA.

\end{itemize}

\subsubsection{Selection and Crossover:}

The selection process involves identifying a pair of chromosomes from the population. Each instance of selection involves choosing an individual chromosome, thus necessitating two runs. We employ tournament selection~\cite{miller1995genetic} as the selection strategy, a popular and efficient method in evolutionary algorithms. This approach randomly selects a subset of $n$ chromosomes from the population and chooses the one with the highest fitness. Each chromosome can be chosen only once per generation, with no repeats allowed.


For the selected pair of chromosomes, the crossover operation exchanges the portion of chromosomes to produce two offspring. Employing the single-point crossover method~\cite{Goodman14}, we randomly select a point on each chromosome for recombination. At these points, the chromosomes undergo recombination, yielding two offspring.
Note that, the sequence of order-serving and charging/discharging actions undergo crossover at the same randomly selected point. We ensure the feasibility of the generated offspring, adding only valid ones back to the population.



\subsubsection{Mutation:}
For the chromosomes in the population pool, the mutation process introduces random changes to simulate genetic mutation. While random changes can diversify the offspring, they are unlikely to result in a valid solution of reasonable quality. To address the EVOP-V2G problem, we customize the mutation strategies as follows:

\begin{itemize}
    \item \textbf{Order Mutation:} The order mutation randomly selects an order-serving action $(v_i) \in C$ from the offspring. Within the time window $[\tau^b_i, \tau^e_i]$ of $v_i$, we select another order $v_j \in V_o$ whose time window falls within the same interval $[\tau^b_i, \tau^e_i]$ to replace $v_i$.
    
    \item \textbf{Charging/Discharging Mutation:}
    The charging mutation randomly selects a charging/discharging action $(v_i, [\tau_{c_{b}}, \tau_{c_{e}}]) \in C$ from the offspring. Subsequently, a new charging/discharging action is generated to replace the original one. This action involve determining both the charging station and amount of charge/discharge, using the charging/discharging insertion phase of Algorithm 1.

    
    \item \textbf{Insertion Mutation:}
     Insertion mutation focuses on introducing the new orders into the offspring $C$. This process involves randomly selecting a position from $C$, followed by the random selection of an order from $V_o$, which is then inserted into the previously determined position.

\end{itemize}

For each mutation iteration, we set the probability of performing each mutation strategy mentioned above to be 0.8. The feasibility of the mutated offspring is then checked, and ignore those offspring that are invalid. Each mutation iteration is executed up to a maximum of 10 times, or until the fitness value of a generated offspring surpasses that of the original parent selected for mutation. This approach ensures that the offspring produced have a higher probability of improving upon the parental traits. 

 


\subsubsection{Replacement:}

In each generation, the algorithm performs \textbf{selection \& crossover} followed by \textbf{mutation} on the feasible chromosomes, according to their respective probabilities, introducing a varying number of new chromosomes into the population. The population is then sorted in descending order based on fitness values, and adjusts the population to the predefined size by removing the chromosomes with the lowest fitness values. This approach ensures that the elite chromosomes of the current population are retained for the next generation. After the replacement phase, the algorithm conclude one generation and starts the next generation.

\begin{example}

Fig.~\ref{fig::ga_fc} illustrates the EA process applied to the toy instances in Fig.~\ref{fig::lns_fc}. Initially, the EA generates $k$ chromosomes using Algorithm~\ref{algo:ga}, each representing a sequence of actions from $V_s$ to $V_d$ that includes order servicing and charging/discharging. For example, chromosome $p_1$ starts at $V_s$, serves $o_2$, $o_3$, and $o_5$, charges at $c_2$, serves $o_7$, discharges at $c_3$, serves $o_9$, and ends at $V_d$.

The evolutionary process begins with crossover: two parents (e.g., $p_1$ and $p_2$) exchange gene segments beyond a random point, producing feasible offspring $c_1$ and $c_2$, highlighted in yellow and green. Then, the mutation process introduces random changes to each chromosome selected by applying one of three mutation types: order, charging/discharging (C/D), or insertion. 
 For instance, $c_1$ undergoes an order mutation replacing $o_5$ with $o_6$, while $c_2$ remains unchanged due to a low mutation probability. Finally, chromosomes are sorted by the fitness value, and only the top $k$ survive. This completes one generation, repeated until the current time $\tau_{curr}$ exceeds the termination time $\tau^d$.

\end{example}

\subsection{Large Neighborhood Search}

\begin{figure*}[bt]
    \includegraphics[width=\linewidth]{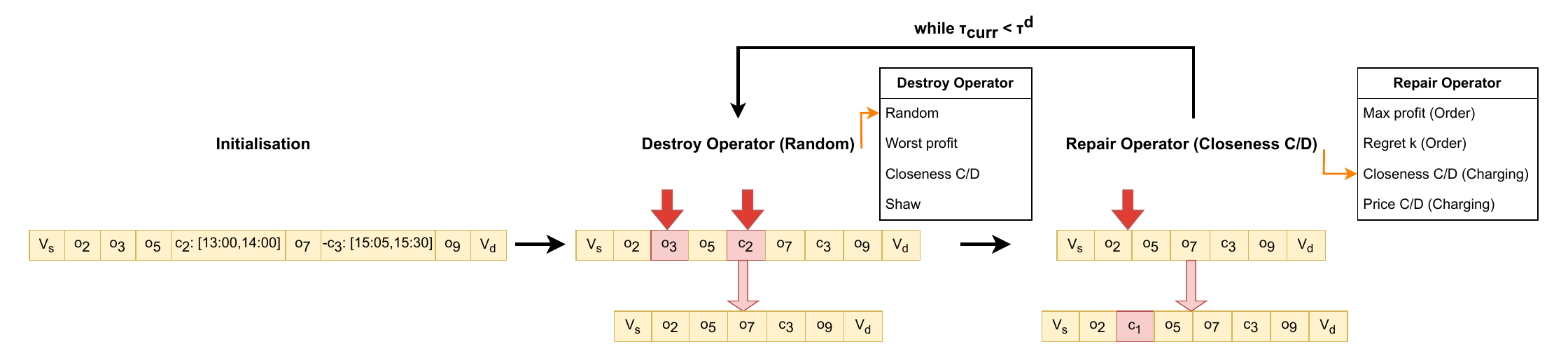}
    \caption{Illustration of the LNS process, showing the stages of initialization, destroy operator, replacement operator, and optimization using MIP.}
    \label{fig::lns_example}
\end{figure*}

Although EA offer a promising solution to the EVOP-V2G problem, it faces two key drawbacks: (i) maintaining a large population of chromosomes leads to slow convergence due to the need to improve multiple solutions simultaneously, and (ii) their reliance on random evolution-inspired operations (e.g., selection, crossover, mutation) results in solution quality being heavily influenced by chance.
To address these issues, we investigate another efficient and successful family of algorithms: Large Neighborhood Search (LNS) \cite{shaw1998using}.
In combinatorial optimization problems, the ``neighbourhood" often refers to a local region or configuration space around a feasible solution. 
Unlike local search, which explores a single, narrowly defined neighborhood, LNS systematically investigates a broader range of neighborhoods. Yet, LNS has been widely adapted for use in the EVRP \cite{erdelic2022goods}, the pickup and delivery problem \cite{ropke2006adaptive}, and other applications.



Following the success of the LNS, we developed an LNS-based algorithm to solve the EVOP-V2G problem. LNS maintains the best solution (i.e., solution with the highest cost) so far and iteratively improves it by exploring the neighborhood around this solution through a ``destroy and repair" process. Fig.\ref{fig::lns_example} shows the overall process of LNS. In broad strokes, the LNS can be described as follows:

\begin{enumerate}
    \item \textbf{Initialization:} Initially, the LNS generates a feasible solution. The algorithm then evaluates the cost of solution and records it as the best solution, $S_{best}$.

    \item \textbf{Destroy Operator:} In each iteration, LNS calls the destroy operator to remove the actions from $S_{best}$, making $S_{best}$ an partial solution, $S_{best}'$.

    \item \textbf{Repair Operator:}
    Taking the partial solution $S_{best}'$, the repair operator explores the neighborhood around $S_{best}$ by repairing $S_{best}'$ into a feasible solution $S$. LNS updates $S_{best}$ to $S$ if the cost of $S$ improves upon the previous $S_{best}$.
\end{enumerate}

Similar to the EA, LNS can run the ``destroy and repair" operations for a certain number of iterations or terminate based on a predefined cut-off time. The current best solution $S_{best}$ is then returned as the final result. Unlike EA, which improves multiple solutions simultaneously at each iteration, LNS focuses on enhancing a single solution at a time. While efficient, achieving high-quality solutions with LNS requires careful design of its destroy and repair operators. Additionally, we introduce optimization to further improve the algorithm.

\subsubsection{Initialization}
To represent a feasible solution, we inherit the same chromosome representation used in EA, where each feasible solution is represented as a set of actions $(v_i, [\tau_{c_{b}}, \tau_{c_{e}}])$. 
Since the objective of LNS is to maximize the amount of profit gained, we reuse the fitness function of EA (i.e., equation~(\ref{eq:fitness})) to evaluate the cost of the solution.
Essentially, the initialization phase only requires generating a feasible solution; the quality of the solution often does not play an important role as the algorithm quickly iterates many iterations to improve the solution.
 To keep the process simple, we initialize $S_{best}$ with actions to visit the source and destination vertices only, without any additional actions in between. In the first iteration, the algorithm skips the destroy operator and calls the repair operator to enhance $S_{best}$ into a high-quality solution.


\subsubsection{Destroy Operator}
Given the current best solution $S_{best}$, the destroy operator deconstructs the solution by removing a set of actions between source and destination. The number of actions removed is a random integer $k$, chosen within minimal and maximal thresholds based on the actions between the source and destination.
The main purpose of the destroy operator is to determine which actions to eliminate, enabling $S_{best}$ to escape local optima. This, in turn, allows the repair operator to reconstruct the solution to a neighbour with higher cost. We designed four different strategies for removing actions from $S_{best}$:

\begin{itemize}
    \item \textbf{Random Removal}: To diversify neighborhood exploration sufficiently, random removal strategies select a random subset of $k$ actions to remove from $S_{best}$.

    \item \textbf{Worst Profit Removal}: 
    The worst profit removal is a heuristic strategy based on the observation that low-profit actions may prevent the solution from achieving higher costs. This strategy removes $k$ actions from $S_{best}$. To evaluate the profit of an action $(v_i, [\tau_{c_{b}}, \tau_{c_{e}}]) \in C$, we consider:

    \begin{itemize}
        \item For serving order $v_i \in V_o$, we take its profit $p_i$.
    
        \item For charging at a station $v_i \in V_c$, we compute the profit is the sum of the differences between the maximum charging price at all stations and the current charging price, i.e., $\sum\limits_{t_k \in [\tau_{c_{b}}, \tau_{c_{e}}]} \left(\max(P^{Ct_k}) - P^{Ct_k}_i \right)$.
    
        \item For discharging at a station $v_i \in V_c$, the profit is the sum of the discharging price, i.e., $\sum\limits_{t_k \in [\tau_{c_{b}}, \tau_{c_{e}}]} \left(P^{Ct_k}_{i}\right)$.
    \end{itemize}

    \item \textbf{Closeness Charging/Discharging Removal}:
    The closeness charging removal method leverages the observation that EV charging/discharging often coincides with nearby orders, e.g., when an EV charges/discharges at a station, it tends to handle nearby orders. This approach aims to reduce these clusters by randomly selecting and removing $k$ actions associated with a charging/discharging event.


    \item \textbf{Shaw Removal}: 
The shaw removal is a heuristic strategy based on the observation that removing correlated actions can lead to better solutions, while removing dissimilar actions often reverts the solution to a similar or worse state.
    Therefore, shaw removal randomly selects an action from $S_{best}$ and removes the $k$ most relevant actions. To measure the relatedness of two actions $(v_i, [\tau_{c_{b}}, \tau_{c_{e}}])$, $(v_j, [\tau_{c_{b}}', \tau_{c_{e}}'])$, we adapt the measurement~\cite{ropke2006adaptive} as follow:
    \begin{equation}
        d(l^d_i,l^d_j) + d(l^p_i,l^p_j) + |\tau_{c_{b}} - \tau_{c_{b}}'| + |\tau_{c_{e}} - \tau_{c_{e}}'|
    \end{equation}
    
    The measurement includes the sum of distances between the pickup and dropoff locations for two actions and their time difference. For an order-serving action $(v_i)$, the action time $[\tau_{c_{b}}, \tau_{c_{e}}]$ is set to its time window $[\tau^b_{i}, \tau^e_{i}]$. For a charging/discharging action $(v_i, [\tau_{c_{b}}, \tau_{c_{e}}])$, the pickup $l^p_i$ and dropoff $l^d_i$ locations are set to the same location as $v_i$.

\end{itemize}

\subsubsection{Repair Operator}

After the destroy operator removes a set of actions from $S_{best}$ to form a partial solution $S_{best}'$, the repair operator then reconstructs the partial solution into a new solution $S$ by adding new actions. The primary objective of the repair operator is to utilize the partial solution $S_{best}'$ to explore the best solution in the neighbourhood nearby. Therefore, the repaired solution must be (i) feasible and (ii) of high quality. Achieving both simultaneously is challenging. To address this, we have designed a simple yet efficient algorithm that iteratively inserts actions to form the new solution 
$S$.


Algorithm~\ref{algo:repair} presents the pseudo-code of our proposed algorithm. Initially, the algorithm takes the partial solution $S_{best}'$ and evaluates the remaining battery level, $B_{last}$, after the last actions of $S_{best}'$ (line~\ref{line::update_battery}). The algorithm then enters a while loop, attempting to insert new actions, which are either order-serving or charging/discharging actions. In each iteration, the algorithm tries to insert both types of actions. If $B_{last}$ is below a threshold $\theta$ of the battery capacity $B$ (line~\ref{line::lower}), the algorithm prioritizes charging/discharging actions to maintain flexibility for order-serving actions, and vice versa (line~\ref{line::insert_begin} - \ref{line::insert_end}). After each insertion, $B_{last}$ is updated. The loop continues as long as at least one insertion is successful and terminates when no more actions can be inserted. Finally, the algorithm returns the repaired solution $S$.
In this paper, we set $\theta$ to 0.15, meaning the algorithm prioritizes charging/discharging actions when the battery level $B_{last}$ falls below 15\%.
Next, we describe different repair strategies for inserting order-serving or charging/discharging actions.

\begin{algorithm2e}[t]
\small
\caption{Repairing the Partial Solution}
    \label{algo:repair}
\setcounter{AlgoLine}{0}
    \SetKwInput{Input}{Input}
    \SetKwInput{Initialization}{Initialization}
    \SetKwInput{Output}{Output}
    \Input{$S_{best}'$: The partial solution returned by destroy operator.}
    \Output{$S$: A feasible solution returned by repair operator.}
    $B_{last} =$ compute the battery level at the last action of $S_{best}'$; \label{line::update_battery}\\
    \While{an action is inserted successfully} 
    {
        \If{$B_{last} <  B \times \theta$ \label{line::lower}}{
            Insert a charging/discharging action in $S_{best}'$; \label{line::insert_begin}\\
            Insert an order-serving action in $S_{best}'$;
        }\Else{
            Insert an order-serving action in $S_{best}'$;\\
            Insert a charging/discharging action in $S_{best}'$; \label{line::insert_end}\\
        }
        $B_{last}$ = update the battery level at the last action of $S_{best}'$;
    }
    \textbf{return} $S = S_{best}'$; 
\end{algorithm2e}

\paragraph{Inserting Order-Serving Actions}
Each order in the order pool $V_o$ can only be served once. To insert an order-serving action, the algorithm scans $V_o$ and skips orders already served in $S_{best}'$. For each remaining order $v_o$, it checks if the EV driver has sufficient time and battery to complete the order while preserving other actions in $S_{best}'$. If no valid orders are available, the algorithm returns a failure and may attempt to insert charging or discharging actions instead. For valid orders, the algorithm selects based on the following strategies:
 

\begin{itemize}
    \item \textbf{Max Profit Insertion:}
For each valid order, it may be possible to insert it at multiple positions in $S_{\text{best}}'$. To determine the order and insertion position, we define the insertion profit $I(v_j)$ for inserting order $v_j$ between $(v_i, [\tau_{c_{b}}, \tau_{c_{e}}])$ and $(v_k, [\tau_{c_{b}}', \tau_{c_{e}}'])$. Let $p_e(v_i, v_j)$ represent the energy cost from $v_i$ to $v_j$, calculated based on energy consumption and the highest charging price. The insertion profit $I(v_j)$ is then computed as follows:

    \begin{equation}
        I(v_j) = p_j  - p_e(v_i,v_j) - p_e(v_j,v_k)
    \end{equation}

    The max profit insertion simply selects the order and position with the highest insertion profit. The intuition here is to maximize the total profit for $S_{best}'$.

    \item \textbf{Regret $k$ Insertion:} 
    The regret-$k$ insertion improves the maximum profit insertion by incorporating a look-ahead mechanism. Let $I_{1}(v_j)$ denote the highest insertion cost for inserting the order $v_j$ into $S_{best}'$. The regret value of inserting order $v_j$ measures the profit lost between inserting $v_j$ at the position with the highest insertion cost and the second highest insertion cost (i.e., $I_{1}(v_j) - I_{2}(v_j)$). This regret value can naturally be extended to consider up to $k$ of the highest insertion costs as follows: 

    \begin{equation}
    R(v_j) = \sum_{n < k}^{n=1} (I_{n}(v_j) - I_{1}(v_j))
    \end{equation}
    
    The regret-$k$ value indicates the importance of inserting the order at the current iteration, which would otherwise result in significant profit loss if inserted later. Hence, the strategy selects the order with the highest regret-$k$ value and inserts it at the position with the highest insertion cost.
    
\end{itemize}

\paragraph{Inserting Charging/Discharging Actions}

Unlike inserting an order-serving action, charging/discharging actions are not bound by strict time constraints, i.e., an EV can charge or discharge whenever it arrives at a station, while an order must be served within its time window. Consequently, considering all possible insertion positions for each charging station is time-consuming. 
To address this, we determine charging/discharging actions by iteratively selecting unique positions from $S_{best}'$ and evaluating potential actions at each charging station.
In each iteration, an EV charges (or discharges) if its battery level is below (or above) the required level for remaining actions, plus a threshold 
$B \times \theta$. Each action is verified to ensure the EV can charge/discharge for at least one timestep and has enough battery to reach and leave the charging station. Similar to EA, we limit the EV to no more than $m$ consecutive charging/discharging actions. If valid actions are identified, the algorithm selects the charging/discharging station based on the following strategies:


\begin{itemize}
    \item \textbf{Closeness Charging/Discharging Insertion:} 
    Among all charging/discharging stations, this strategy selects those where the charging station is closest to the adjacent actions in $S_{best}'$. The intuition is to choose nearby charging stations, making $S_{best}'$ more flexible for subsequent order-serving or charging/discharging activities.
    
    \item \textbf{Price Charging/Discharging Insertion:} 
    For each charging/discharging station, we consider the variable price of station throughout the day, assuming the EV can charge/discharge for the maximum available timesteps. A lower price is preferred for charging, while a higher price is better for discharging.
    The price charging insertion selects the charging/discharging station with the best prices, aiming to maximize the profit collected in $S_{best}'$.
\end{itemize}

Once the station is selected, we randomly chooses a number of timesteps between 1 and the maximum available at the insertion position and return the charging/discharging actions accordingly. If no valid action is found, it checks the next random position and returns a failure after evaluating all positions.


\begin{example}

Fig.\ref{fig::lns_example} shows the main steps of the LNS algorithm applied to the toy instance in Fig.\ref{fig::lns_fc}. The process begins by repairing the trivial initial solution $\{V_s, V_d\}$. Then, the algorithm iteratively applies a destroy-and-repair process to improve the solution. Four destroy strategies are available: random, worst profit, closeness C/D, and Shaw. In each iteration, one strategy removes part of the current best solution. For instance, the random strategy removes $o_3$ and $c_2$. The repair phase then reconstructs the solution using order-serving (max profit, regret $k$) and C/D (closeness, price) operators. In this example, the closeness C/D repair adds $c_1$ at position 2. The process repeats until the current time $\tau_{curr}$ exceeds the time limit $\tau^d$.

\end{example}

\subsubsection{Optimization}
So far, we have introduced various strategies for the destroy and repair operator, each incorporating a greedy approach that ensures efficiency. 
While each iteration is fast, achieving a high-quality solution requires guidance to select a suitable strategy.
To enhance LNS, we introduce a set of optimizations:

\paragraph{Adaptive LNS~\cite{ropke2006adaptive} (ALNS)} ALNS is an enhanced version of LNS that employs multiple strategies, tracking their success in improving the current solution and using the most promising strategy to choose the next neighborhood.
    To implement the strategies selection, assume we have a set of strategies $\mathcal{S}$. Initially, each strategy is assigned a weight value $w_i = 1$. In each iteration, strategies are selected using the roulette wheel selection method~\cite{roulette_wheel}, where a strategy $i$ is chosen with a probability of $\frac{w_i}{\sum w_j}$ for $j \in \mathcal{S}$. After an iteration, the weight of the selected strategy $i$ is updated based on the improvement in the solution. Specifically, $w_i$ is adjusted as follows:

    \begin{equation}
        \alpha \times max\{S - S_{best}, 0\} + (1 - \alpha)w_i
    \end{equation}

The parameter $\alpha$ is a user-defined reaction factor that determines the rate at which the weights adjust in response to variations in the relative success of enhancing the current solution. In our experiments, we set $\alpha$ to 0.01. Note that the strategy selection for the destroy and repair operators is conducted independently. For the repair operator, we simplify the selection process by considering permutations of strategies for inserting the order-serving and charging/discharging actions.

\paragraph{Introducing Randomness to Strategies} 

LNS uses various strategies for selecting destruction or repair actions, typically through a greedy approach: (i) rank actions based on a single measurement and (ii) select the top one or $k$ actions to remove or insert. While different strategies may be employed in each iteration, relying on a single measurement can lead to local optima. To address this, we introduce randomness into the action selection process.
Assuming we have $N$ actions sorted by measurement, instead of selecting the top action, we choose an action at the position of $N$ based on the following criteria:

\begin{equation}
    r \in [0,1] : r^m \times N
\end{equation}

The $m$ value represents the degree of freedom. Higher $m$ values indicate a greater likelihood of selecting an action from the top few options. To ensure the algorithm maintains a high probability of selecting the best action based on the measurement, we set $m = 5$ in our experiments.

\paragraph{Improving Solution Using MIP Solver} 
Recall that when inserting a charging/discharging action, the repair operator determines the amount by randomly assigning a number of time steps to each action. This method can frequently result in suboptimal charging/discharging solutions. i.e., charging/discharging more at one station may result an unnecessary visit to a subsequent station, thereby reducing overall profit.
To address this issue, we reuse the MIP model proposed in Section~\ref{sec:MIP} to further optimize the charging/discharging actions.

While the overall MIP model remains the same, we modify the input graph. Specifically, using the repaired solution $S$, we build an input graph containing a single path that starts from the source and ends at the destination, visiting orders and charging stations in between. Each order vertex must be visited once, but for each charging station, we add an edge between adjacent vertices to allow the model to skip the charging station. Given the modified input graph, the MIP model only needs to reason about the charging/discharging period and the selection of the station, thereby the model often find the solution fast. To further improve efficiency, we set the suboptimality gap to 5\% and allow the solver to terminate upon finding a solution within 5\% of the optimal. This prevents the solver from running excessively long at each iteration. 
\section{Evaluation and Results}
\label{sec::Evaluation}

\subsection{Settings}


To evaluate our proposed algorithms, we use a dataset created by integrating real-world ride-sharing trip statistics, EV specifications, and ride-sharing fare prices. The EVOP-V2G algorithms rely on two main inputs: a set of orders and a set of charging stations. 
First, we describe how we obtained the order candidates and charging station candidates from real-world datasets. 
These candidates are later used in the experiments for evaluating our proposed algorithms.



\textbf{Order Candidates:}
We use ride-sharing orders from a public dataset~\cite{dataset}, which contains real-world trip records in and around Melbourne, Australia. Among the three available sizes, we chose the largest, with 68,625 orders. Each order includes a pickup location $l_i^p$, drop-off location $l_i^d$, travel distance $d_i$, travel time $t_i$, and a time window $[\tau_i^b, \tau_i^e]$. Focusing on the Melbourne metropolitan area, we exclude orders where both pickup and drop-off lie outside this region, resulting in 64,487 valid orders.
The profit $p_i$ for each order is based on a fare structure: a 2.75 base fare, 1.49/km distance fare, and 0.39/min time fare, with a 30\% deduction reflecting typical ride-sharing service fees~\cite{ridesharingAus2023,uberEarningsCalculator2023}. All amounts are in AUD.

To simulate realistic scenarios, we vary order selection based on three criteria: (i) Bounding Box: filtering orders using 10\%, 40\%, 70\%, and 100\% of the Melbourne metropolitan area; (ii) Ride Length: grouping trips into 5–10 km, 10–25 km, and over 25 km; and (iii) Time Period: selecting orders from 2-hour (9–11 am), 5-hour (9 am–2 pm), and 8-hour (9 am–5 pm) windows. By default, we set the bounding box to 40\%, ride length to 10–25 km, and time period to 8 hours, varying one criterion at a time while keeping the others fixed.

\begin{figure*}[t]
  \begin{subfigure}{.34\textwidth}
    \includegraphics[width=\linewidth]{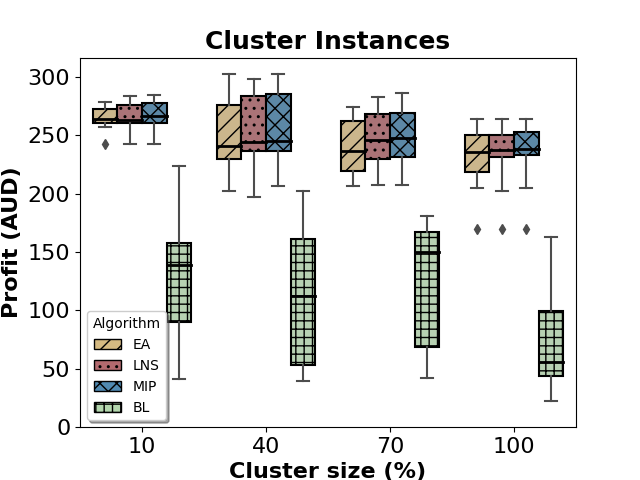}
    \caption{}
    \label{fig::small_cluster}
  \end{subfigure}%
  \begin{subfigure}{.34\textwidth}
    \includegraphics[width=\linewidth]{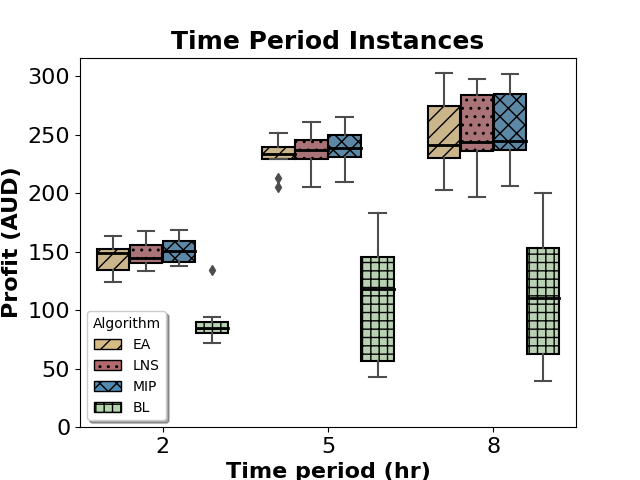}
    \caption{}
    \label{fig::small_tw}
  \end{subfigure}
  \begin{subfigure}{.34\textwidth}
    \includegraphics[width=\linewidth]{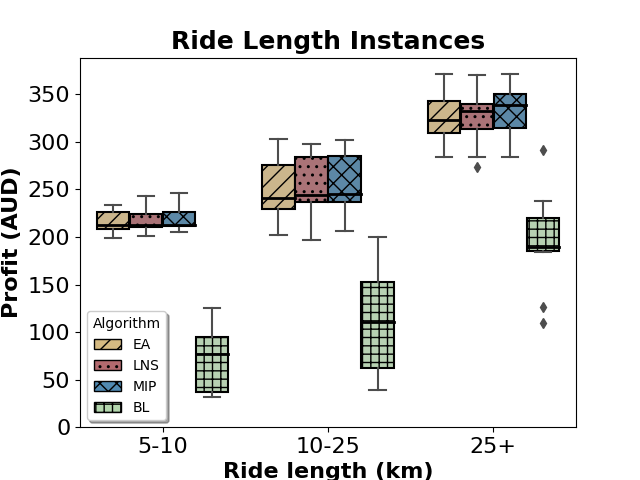}
    \caption{}
    \label{fig::small_length}
  \end{subfigure}
  \caption{Small scaled instances. The x-axis represents the variation of parameters for each criterion. Fig.~\ref{fig::small_cluster} shows the cluster bounding box as percentage \%, Fig.~\ref{fig::small_tw} illustrates the time window in hours hr, and Fig.~\ref{fig::small_length} displays the ride length range in kilometers (km). The y-axis shows the overall profit in AUD. Each boxplot for each competitor represents the distribution of profit generated over the 10 repeats of same parameter. The box represents the interquartile (IQR) range showing 50\% of the profit, with the line inside marking the median. Whiskers extend to the smallest and largest values within 1.5 times the IQR from the first and third quartiles, respectively. The diamond-shaped dots denote outliers that is outside the range. }
  \label{fig::small}
\end{figure*}

\textbf{Charging Station Candidates:} 
We collected data from PlugShare\footnote{\url{https://www.plugshare.com/}}, initially identifying 440 EV charging stations across Australia. After removing 171 stations lacking charging rate data and those outside the Melbourne metropolitan area, 70 stations remained. 
For discharging price of each station, we use the pricing scheme provided by Energy Australia's feed-in tariff (FOU) price \footnote{\url{https://www.energyaustralia.com.au/home/solar/feed-in-tariffs}}. As of 2024, this scheme specifies three time periods: peak (4pm - 9pm), shoulder (9pm - 10am and 2pm - 4pm), and off-peak (10am - 2pm). The corresponding prices are \$0.117 (peak), \$0.061 (shoulder), and \$0.043 (off-peak).

 We assume that the source and destination (e.g., home of the EV owner) are equipped with charging infrastructure, with a charging rate of 7 kw/h. The charging and discharging prices for the source and destination follow a standard two-period time-of-use (TOU) tariff. During the peak period (3pm - 9pm), the price is \$0.412 per kWh. During the off-peak period, covering all other times, the price is \$0.2665 per kWh~\cite{VictorianDefaultOffer2023}.

\textbf{EV Specifications:}
In our experiments, we utilize the specifications of the Tesla Model S for our EV setting, with a battery capacity of 70 kWh and a driving range of 400 km. The driving efficiency $\gamma$ can be calculated as the ratio of battery capacity to driving range which is 0.175 kWh/km in our experiments. Please note that these EV settings are taken in as parameters and can be easily adjusted to match other EV specifications. The working hours for the EV owner are set to the standard normal working hours, with $\tau_w^s$ set to 9 AM and $\tau_w^e$ set to 5 PM. We set the initial battery level $B^s$ at source as 100\% and set minimal battery required $B^d$ at destination as 0\%.

\textbf{Competitor and Implementation:} 
Since there is no existing work that addresses the same problem, we propose a simple local search algorithm as a baseline approach (BL). This baseline approach first sorts the available orders into time windows and then greedily selects the closest time window. From this selected time window, it randomly inserts order-serving actions until the current actions exceed the working hours limit, at which point the vehicle returns to its destination. BL handles charging and discharging in a greedy manner. When the battery level is greater than 20\%, BL preferentially inserts order-serving actions. When the battery level is less than 20\%, BL inserts charging actions. The charging stations are selected in a similar approach to EA, the charging station candidates compute a score based on distance and charging price, then the BL algorithm randomly selects out of the top 5 candidates.  When charging, we assume that the charging continues until the EV is charged to full capacity. Discharging actions are only inserted when the EV returns to its destination, at which point we assume that the EV discharges all remaining energy until it reaches the battery threshold. 
The BL executes a while loop, until the first feasible solution is found and returned.

For our proposed algorithm, including MIP, EA, and LNS, as well as the baseline algorithm, all implementations are in C++. We solve the MIP model using the off-the-shelf solver Gurobi (version 10.0.3)\footnote{\url{https://www.gurobi.com}}. For the EA algorithm, we set the population pool to 2000. In the initial stage of the experiments, we also varied the population pool size to other values, but this did not significantly affect the results, and in some cases, worsened them. We conduct all experiments on an Apple M1 Pro with 16GB RAM. For all algorithms, except for MIP, we set a runtime limit of 60 seconds for small-scale instances and 300 seconds for large-scale instances. After these time limits, we return the best results obtained by each algorithm. We allow MIP to run until an optimal solution is found (which takes up to 27 hours for some instances). For reproducibility, our implementation is available online~\footnote{\url{https://will-publish-after-acceptance.com}}.

\subsection{Experiment Results}

\subsubsection{Small-Scale Instances}
\textbf{Instance:} 
As mentioned in Section 4, the MIP model encounters scalability issues, making it impractical to execute on larger instances. Therefore, to compare its performance against other algorithms, we create the small-scale instances. For each instance, we set the source and destination to the same location (i.e., the home of the EV owner) and designate three charging locations: one at the home of the EV owner and the other two selected randomly from the charging station candidates. In each experiment, we generate 10 instances, where each instance consists of 30 randomly selected orders from the corresponding order candidates. We report the results of these 10 instances.

\begin{figure*}[t]
\begin{subfigure}{.34\textwidth}
    \includegraphics[width=\linewidth]{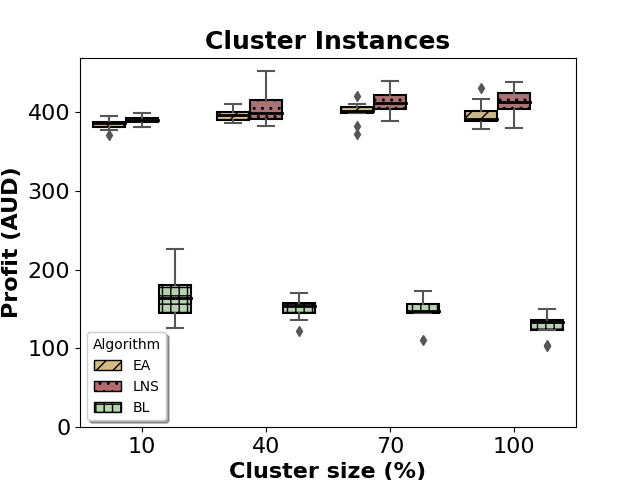}
    \caption{}
    \label{fig::default_cluster}
  \end{subfigure}%
\begin{subfigure}{.34\textwidth}
    \includegraphics[width=\linewidth]{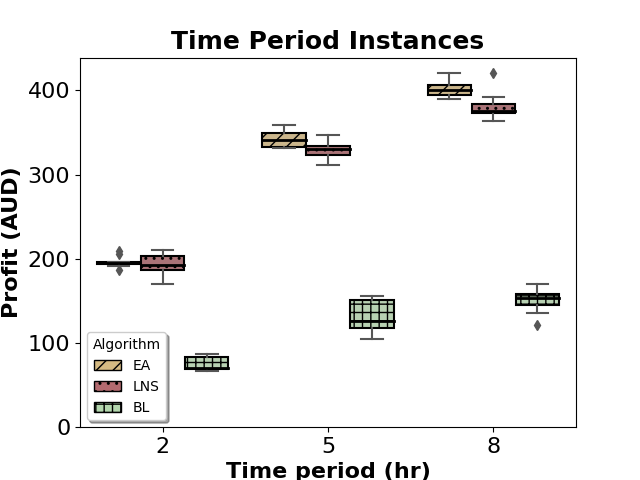}
    \caption{}
    \label{fig::default_tw}
  \end{subfigure}%
\begin{subfigure}{.34\textwidth}
    \includegraphics[width=\linewidth]{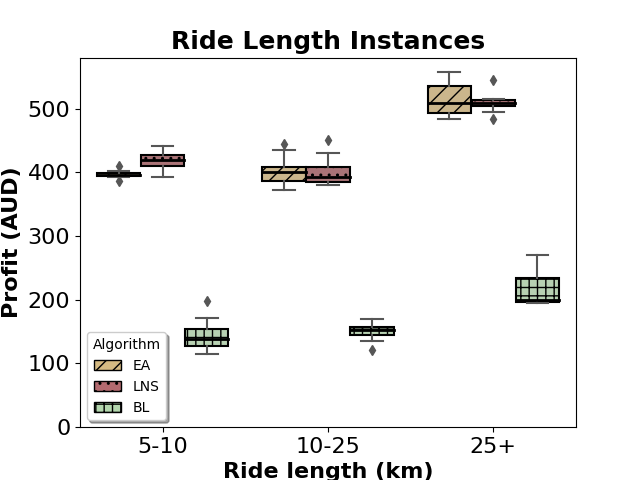}
    \caption{}
    \label{fig::default_length}
  \end{subfigure}
\caption{Large-scale instances. The x-axis represents the variation of parameters for each criterion. Fig.~\ref{fig::default_cluster} shows the  bounding box size as percentage \% of the Metropolitan area, Fig.~\ref{fig::default_tw} illustrates the time window in hours (hr), and Fig.~\ref{fig::default_length} displays the ride length range in kilometers (km). The y-axis shows the overall profit in AUD. Each boxplot for each competitor represents the distribution of profit generated over the 10 repeats of same parameter. The box represents the interquartile (IQR) range showing 50\% of the profit, with the line inside marking the median. Whiskers extend to the smallest and largest values within 1.5 times the IQR from the first and third quartiles, respectively. The diamond-shaped dots denote outliers that is outside the range.}
  \label{fig::default}
\end{figure*}

\textbf{Results:} 
Fig.~\ref{fig::small} presents 
the boxplots for our proposed algorithms (MIP, LNS and EA) and BL on the small-scale instances. Each boxplot illustrates the distribution of profit generated by an algorithm over the 10 different instances. The box in each plot represents the interquartile range (IQR), encapsulating the middle 50\% of the profit data. The line within the box marks the median value. The whiskers extend to the smallest and largest values within 1.5 times the IQR from the first and third quartiles, respectively. Observations that fall outside this range are considered outliers and are denoted by diamond-shaped dots.
In our observations of the solution outputs from EA and LNS, we found that these solutions often included multiple revisits to a single charging station due to its low or free charging costs. To ensure fair comparison, we determined the maximum number of station revisits across EA and LNS (up to 3) and set this as a constraint in the MIP model. Due to computational limits (solving time more than 48 hours), we restricted revisits to 2, which still enabled MIP to outperform EA and LNS on all instances. MIP computation times ranged from 5 seconds to 27 hours. 
Overall, our proposed algorithms MIP, LNS, and EA significantly outperform the baseline algorithm (BL), achieving more than two times higher profit. Among these, the MIP approach consistently delivers highest overall profits compared to EA and LNS across all instances. However, EA and LNS remain highly competitive, achieving near-optimal solution quality relative to MIP.

Fig.~\ref{fig::small_cluster} illustrates the impact of the bounding box size with settings of 10\%, 40\%, 70\%, and 100\%. 
As the size of the bounding box increases, the overall profit collected by EV drivers remains similar but slightly decreases when the bounding box size reaches 100\% for our proposed algorithms. This occurs because a larger bounding box implies that the orders are spread across a larger area which requires EVs to travel longer distances between orders, leading to more wasted time. 
Fig.~\ref{fig::small_tw} demonstrates the effect of increasing time periods  from 2 to 5 and 8 hours. With longer time periods, orders are more temporally dispersed, allowing EVs to serve more requests within the same working hours, thereby increasing total profit.
Finally, Fig.~\ref{fig::small_length} examines the influence of ride lengths ranging from 5-10 km, 10-25 km, to more than 25 km. As ride length increases, so does the total profit, largely due to higher fares per order. Despite a relatively consistent number of completed orders across methods, longer rides contribute more to profit. However, for rides over 25 km, profit variability is higher; missed high-fare orders due to suboptimal sequencing can lead to lower performance compared to mid-range rides. This highlights the importance of effective order selection and routing, particularly in high-value scenarios.

\subsubsection{Large-Scale Instances}

\textbf{Instances:}
To further assess the performance of LNS and EA relative to each other and the baseline, we evaluate them on larger instances. Similar with the small-scale experiments, we generate 10 instances for each order candidate variation. Each large-scale instance includes 900 randomly selected orders and all 70 charging stations. The number 900 reflects the size of the filtered dataset under the most restrictive criteria. For each instance, the EV’s source and destination are identical and randomly selected within the bounding box of the order set.


\textbf{Results:}
 Fig.~\ref{fig::default} shows the results of our algorithms (LNS and EA) and BL on large-scaled instances. From the figure, it is evident that both LNS and EA outperform the BL, confirming the significant advantage of utilizing our proposed algorithm in real-world scenarios. Compared to EA, LNS achieves higher profits for most of the instances. This further demonstrates the advantage of LNS, where the higher profit is primarily due to LNS's ability to smartly select destroy or repair strategies to efficiently escape local optima. In contrast, EA simulates a bio-inspired evolutionary process. While it is relatively simple compared to the more sophisticated strategies of LNS, the improvement in solution quality often depends on the randomness introduced by the selection, crossover, and mutation processes. 
However, as the experiments demonstrate in Fig.~\ref{fig::default}, the randomization in EA can be particularly advantageous for certain problem instances (such as longer time periods) where diverse exploration of the solution space is required. In instances where EA outperforms LNS, this advantage is due to EA's ability to leverage randomization through its evolutionary processes, such as crossover and mutation, to explore a wider variety of solutions and avoid getting trapped in local optima. This broad exploration enables EA to discover higher-quality solutions that LNS, which tends to focus on local improvements, may miss.

Regarding the different variations of order candidates, the overall trends observed in large-scale instances are similar to those in small-scale instances. However, in large-scale instances, the total profit collected by EV drivers increases, and the data points in the boxplot are more densely clustered. This is because large-scale instances are characterized by a greater number of orders and charging stations, which helps in finding better, higher-profit solutions.


\begin{figure}[t]
\begin{subfigure}{.25\textwidth}
    \includegraphics[width=\columnwidth]{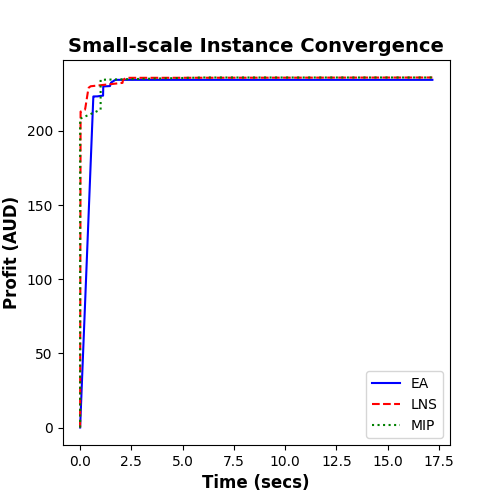}
    \caption{Small-Scale Instance }
    \label{fig::small_conv}
  \end{subfigure}%
\begin{subfigure}{.25\textwidth}
    \includegraphics[width=\columnwidth]{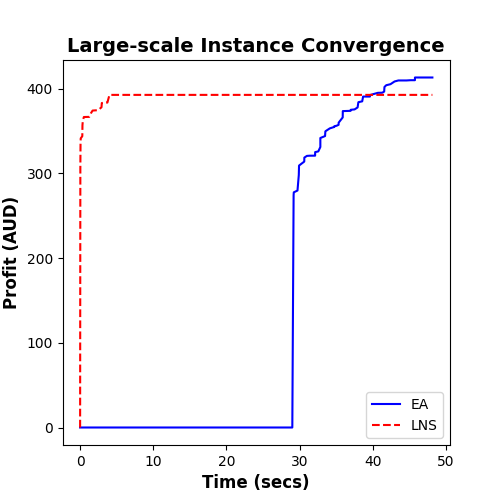}
    \caption{Large-Scale Instance}
    \label{fig::large_convergence}
  \end{subfigure}%
\caption{Convergence graphs for small and large instances comparing our three approaches. The x-axis represents the time taken to discover an improved solution (in seconds), while the y-axis indicates the profit in AUD. The x-axis extends to the final timestamp at which any approach identified a better solution.}
      \label{fig::convergence}
\end{figure}

\subsubsection{The convergence rate of different algorithms}

To further evaluate the performance of the different algorithms, we randomly selected one instance from both the small-scale and large-scale instances and plotted their convergence rates. Fig.~\ref{fig::small_conv} shows the convergence rates of EA, LNS, and MIP on a small-scale instance. With each new solution found, we plotted the profit of the solution and the time in seconds required to find that solution. The plot ends when all the methods have found their best solutions. As shown in the figure, for the small-scale instance, the problem is relatively easy to solve, and all algorithms converge quickly. Even the MIP approach finds a solution of reasonably good quality rapidly.

For the large-scale instance, Fig.~\ref{fig::large_convergence} presents the results for LNS and EA under the same settings. As the problem becomes more complex, LNS quickly finds the first solution and converges to the best solution within 2 seconds. On the other hand, EA takes much longer to initialize the population, requiring approximately 29 seconds to find the first solution. Nonetheless, EA converges much more slowly than LNS, taking up to 50 seconds to reach the best solution.

\subsubsection{Effect of varying charging rate, price, and order fares}

\begin{figure*}[t]
\begin{subfigure}{.34\textwidth}
    \includegraphics[width=\linewidth]{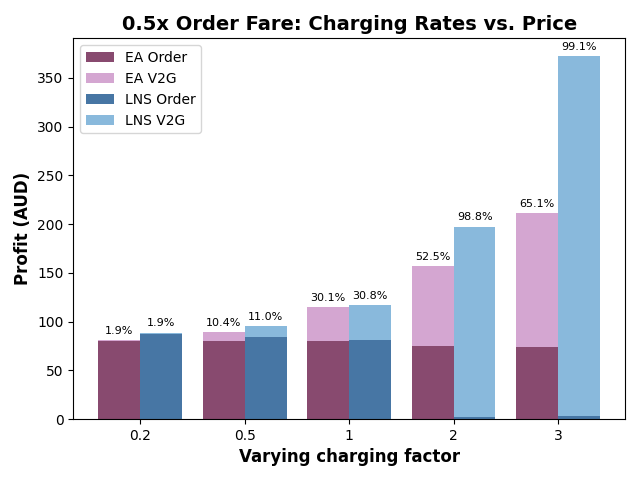}
    \caption{0.5 factor decrease in order fare}
    \label{fig::decrease_order}
  \end{subfigure}%
\begin{subfigure}{.34\textwidth}
    \includegraphics[width=\linewidth]{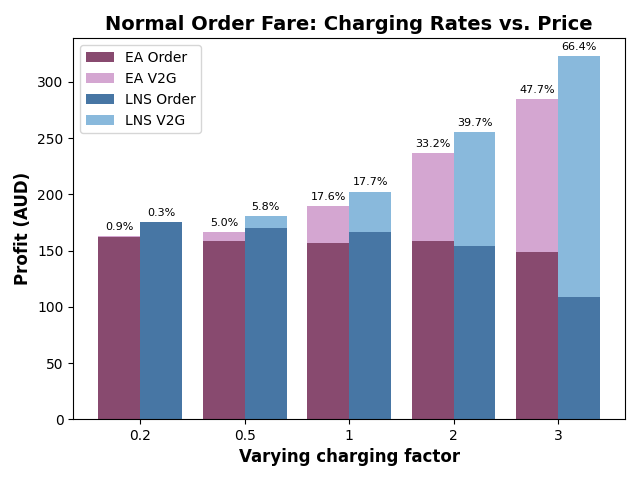}
    \caption{Normal order fare}
    \label{fig::normal_order}
  \end{subfigure}%
\begin{subfigure}{.34\textwidth}
    \includegraphics[width=\linewidth]{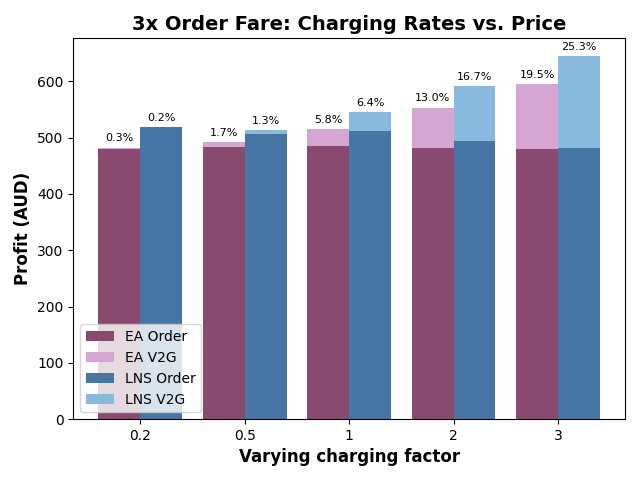}
    \caption{3 factor increase in order fare}
    \label{fig::increase_order}
  \end{subfigure}%
\caption{Variation of charging rates, prices, and order fares for 2-hour time period instances. The x-axis represents the varying factors of both charging rate and price, while the y-axis shows the profit generated in AUD. Fig.~\ref{fig::decrease_order} shows the results when order fares are decreased by a factor of 0.5, Fig.~\ref{fig::normal_order} shows the result of normal order fares, and Fig.~\ref{fig::increase_order} shows the results when order fares are increased by a factor of 3. The lighter shades of each algorithm represent the profit generated from discharging, with annotations on top of the bars indicating the percentage of discharging incentive out of the overall profit.}
  \label{fig::vary_order}
\end{figure*}

\textbf{Instances:} 
With the increasing adoption of electric vehicles (EVs) and advancements in fast-charging technologies, charging rates and prices are expected to evolve significantly. To account for these developments, we examine the impact of varying charging rates and prices within the EVOP-V2G problem.
Specifically, we consider the large-scale instances where the time window for the orders is 2 hours. This choice is based on observations from previous experiments, which indicated that a shorter operational order time-frame for the EVs provides greater opportunities for charging/discharging activities outside this period.
In our experiments, we vary the charging rates and prices at each station by scaling factors of 0.2, 0.5, 1, 2, and 3. Additionally, we modify the fare of each randomly selected order by increasing it by a factor of 3 and decreasing it by a factor of 0.5, respectively. We then evaluate the combined effect of these variations on the performance of algorithm.

\textbf{Results:}
Fig.~\ref{fig::vary_order} presents bar plots of the total profit achieved by each algorithm for varying factors of charging rate and price. Fig.~\ref{fig::increase_order} and Fig.~\ref{fig::decrease_order} shows the effect of varying the order fares in addition to varying the charging rate and price. We chose to increase the charging rate and price simultaneously not only because charging rate and price are usually correlated but also to better illustrate the trend observed when analyzing them separately. We run LNS and EA for each setting and display the average values over 10 instances. For each bar, the lighter-colored area denotes the V2G profit (i.e., by buying and selling energy) by each algorithm, and the annotation on top indicates the percentage of the total profit earned by V2G. 

\underline{\textit{Charging rate and price: }}
At the charging and order default settings (factor of 1) which are based on real-world data, V2G profit is about 20\% of the total profit earned as shown in fig.~\ref{fig::normal_order}. This demonstrates that V2G has the potential to significantly increase the profits.
As expected, when the charging rate and price increase, the contribution of V2G profit to the total profit becomes more pronounced, e.g., when price and charging rate are increased threefold, the percentage of V2G profit rises to 47.7\% for EA and 66.4\% for LNS. As the charging rate and price increase, V2G becomes more profitable and, as a result, the profit earned by orders is reduced as the algorithms start to prefer V2G over some of the orders.

While the profit earned by doing orders shows a downward trend for both EA and LNS, the effect is more significant on LNS. The overall profit for both EA and LNS grows due to the increase in charging rates and prices when the charging rate and price increases. This demonstrates that, as the charging rates and perhaps prices increase in future, the advantages of V2G are expected to be more obvious. For decreasing charging rates and prices (i.e., when the increase factors are 0.5 and 0.2), the trend shows a relative decrease in V2G profit compared to order profit. As expected, when charging rates and prices go down, there is less motivation to discharge, which affects the overall profit. Consequently, the main source of overall profit shifts to the profit made from orders.  
Comparing LNS and EA, it is observed that LNS exhibits better performance in making discharging decisions, thereby achieving higher profits. This is because LNS utilizes the MIP model to optimize charging/discharging actions in each iteration. In contrast, EA initially applies a two-phase algorithm that preferentially inserts order actions first, followed by charging/discharging actions. The more advanced charging/discharging strategies can only be introduced through the mutation process in EA.

\underline{\textit{Order fares: }}
Fig.~\ref{fig::decrease_order} illustrates the results when order fares decrease by 0.5 under different charging rates and prices. When order fares decrease, total profit reduces significantly, especially at lower charging rates and prices. However, as charging rates and prices increase, V2G profit becomes dominant, particularly for LNS, where V2G profit reaches up to 99.1\%. 

Fig.~\ref{fig::increase_order} shows the results when order fares increase by a factor of 3. In this scenario, total profit increases significantly, primarily driven by order profits. Although V2G profit appears marginal under most settings, it still contributes 19.5\% to 25.3\% of the total profit for EA and LNS when charging rates and prices are tripled. This demonstrates that both EA and LNS are capable of optimizing profits from orders effectively, with comparable solution quality.

\section{Conclusion}
\label{sec::Conclusion}

In this paper, we introduce the Electric Vehicle Orienteering Problem with Vehicle-to-Grid (EVOP-V2G) to help commercial EV drivers optimize their daily profits. We propose three distinct algorithms to solve the EVOP-V2G: the optimal Mixed-Integer Programming (MIP) algorithm, and two suboptimal algorithms, Large Neighborhood Search (LNS) and Evolutionary Algorithm (EA). Experimental results demonstrate that our proposed algorithms significantly outperform the baseline algorithm, achieving solutions that yield twice the profit. In small-scale instances, our suboptimal algorithms EA and LNS attain near-optimal solutions compared to MIP, while in large-scale instances, LNS outperforms EA by providing better solutions. Additionally, we simulate future scenarios by varying order profit, and charging/discharging rates and prices. The experiments show that with increases in charging/discharging rates and prices or decreases in order profit, our proposed approaches can generate solutions with higher profits, indicating potential benefits for the future. 

\section*{CRediT authorship contribution statement}

\textbf{Jinchun Du}: Conceptualization, Data curation, Methodology, Software, Validation, Formal analysis, Visualization, Writing-original draft, Writing-review \& editing.
\textbf{Bojie Shen}: Conceptualization, Methodology, Software, Writing-review \& editing.
\textbf{Muhammad Aamir Cheema}: Conceptualization, Supervision, Methodology, Project Administration, Writing-review \& editing.
\textbf{Adel N.Toosi}: Conceptualization, Supervision, Methodology, Writing-review \& editing.

\section*{Data availability}

Data will be made available on request.
\section*{Declaration of generative AI and AI-assisted technologies in the writing process.}

During the preparation of this work the author(s) used ChatGPT in order to assist with grammatical checks and presentation improvements. After using this tool/service, the author(s) reviewed and edited the content as needed and take(s) full responsibility for the content of the published article.

\label{}





\bibliographystyle{elsarticle-num} 
\bibliography{reference}

\end{document}